\documentclass[5p]{elsarticle}

\usepackage{lineno,hyperref}
\usepackage{hyperref}
\usepackage{algorithm}
\usepackage{algpseudocode}
\usepackage{graphicx}
\usepackage{subfig}
\usepackage{colortbl}
\usepackage{amsmath}
\usepackage{siunitx}
\usepackage{multirow}
\usepackage{pdflscape}
\usepackage{tabularx}
\usepackage{amsmath}
\usepackage{url}
\definecolor{gray}{rgb}{0.86,0.86,0.86}
\newcommand{\bhline}[1]{\noalign{\hrule height #1}}
\newcommand{\bvline}[1]{\vrule width #1}

\journal{Applied Soft Computing}

\begin{document}

\begin{frontmatter}

\title{A Fresh Approach to Evaluate Performance in Distributed Parallel Genetic Algorithms}

\author[1]{Tomohiro Harada\corref{mycorrespondingauthor}}
\ead{harada@tmu.ac.jp}
\ead[url]{https://www.comp.sd.tmu.ac.jp/tomohiro-harada/index_en.html}
\author[2]{Enrique Alba}
\ead{eat@lcc.uma.es}
\author[2]{Gabriel Luque}
\ead{gabriel@lcc.uma.es}
\cortext[mycorrespondingauthor]{Corresponding author}
\address[1]{Faculty of System Design, Tokyo Metropolitan University, Tokyo, Japan}
\address[2]{ITIS Software, University of M\'{a}laga, M\'{a}laga, Spain}

\begin{abstract}
This work proposes a novel approach to evaluate and analyze the behavior of multi-population parallel genetic algorithms (PGAs) when running on a cluster of multi-core processors. In particular, we deeply study their numerical and computational behavior by proposing a mathematical model representing the observed performance curves. In them, we discuss the emerging mathematical descriptions of PGA performance instead of, e.g., individual isolated results subject to visual inspection, for a better understanding of the effects of the number of cores used (scalability), their migration policy (the migration gap, in this paper), and the features of the solved problem (type of encoding and problem size). The conclusions based on the real figures and the numerical models fitting them represent a fresh way of understanding their speed-up, running time, and numerical effort, allowing a comparison based on a few meaningful numeric parameters. This represents a set of conclusions beyond the usual textual lessons found in past works on PGAs. It can be used as an estimation tool for the future performance of the algorithms and a way of finding out their limitations.
\end{abstract}


\end{frontmatter}


\section{Introduction}

Most optimization tasks found in Industry and real-world applications impose several constraints that usually do not allow the utilization of exact methods~\cite{Alba2009book,Talbi2009}. Genetic Algorithms (GAs) are well-known population-based techniques for providing accurate solutions to such problems~\cite{kramer2017genetic}. GAs make use of a population of solutions, with the initial population being randomly generated (or created with a greedy algorithm) and then enhanced through an iterative process. At each generation of the process, the whole population (or a part of it) is replaced by newly generated individuals (often the best ones)~\cite{Blum2003}. 

On the one hand, the use of a population of solutions along with genetic operations makes GAs time-consuming in order to reach an acceptable solution, but, on the other hand, its implicit decentralized nature allows to parallelize them as a promising approach for overcoming this drawback. Several parallel models have been then proposed to build parallel GAs (PGAs) \cite{alba2005parallel, talbi2006parallel,harada2020parallel}. The most popular type of algorithm is the island-based model, also known as the distributed model~\cite{Tomassini2006}. In this scheme, the whole population is divided into subpopulations (islands) distributed over different processors. These islands run a sequential GA, and after a predefined interval (migration gap), they communicate in order to exchange some search knowledge, e.g., individuals or search parameters. This communication or migration process \cite{Alba1999,alba2013parallel} usually involves sending copies of a few selected individuals from one island to another in a given topology of migration (e.g., a uni-directional ring). Then, according to predefined acceptance criteria (e.g., always, or just if they have better quality than some other existing ones), the migrated individuals are integrated into the local subpopulation.

The parallel execution of a PGA reduces the total run time by performing more steps per unit time and could even allow for an improvement in the quality of the found solutions since structuring the population in smaller subpopulations leads to new and more efficient ways of exploring the search space~\cite{alba2005parallel}. Most of the modern computing devices such as laptops and workstations are equipped with multi-core processing units. They are also a compact and easy way to develop high-performance computing (HPC) in any present lab and company. However, they are still not well known as the classical plain cluster of CPUs linked by Ethernet for running PGAs. 
This paper analyzes the search performance of island-based PGAs running in a cluster of multi-core computers from the points of view of their numerical effort and computational effort. In particular, we comprehensively analyze the impact of the number of cores used, part of their migration strategy, and the type of search landscape of the targeted problem.

Our research questions are these:
\begin{description}
\item[RQ1:] How do the number of cores, the problem size, and the migration gap affect the search performance of PGAs?
\item[RQ2:] Can we get beyond the mere visual inspection of graphs/tables to analyze speed-up in PGAs and other metrics?
\item[RQ3:] How do the computational and numerical efforts relate to problem size and discrete nature?
\end{description}

To answer these research questions, we conduct experiments using a cluster of multiprocessors, a commodity MIMD (Multiple Instruction, Multiple Data) architecture. As a PGA method, this paper uses a distributed island-based steady-state GA (dssGA). Our experiments focus on the two NP-hard problems: the P-PEAKS problem, using a binary encoding, and the vehicle routing problem (VRP), which uses permutations. This choice is easy to justify since they represent two very important families of problems in combinatorial optimization. To answer RQ1, we solve these two problems with different sizes (dimensions) while also studying the migration gap and the different cores used (a high number of combinations and independent runs needed). Although a migration strategy involves making many decisions (gap, selection policy, and distributed topology), this paper focuses on the migration gap as a first approach to test our fresh understanding of their performance under a wide set of configurations. Indeed, the migration gap highly impacts the diversity of the island-based PGA. For RQ2, we devise a mathematical model that enables a higher level and quantitative analysis beyond the usual comparison of figures in tables/graphs found in the literature. We then used all that for a final study of PGAs to better understand their computational and numerical efforts (RQ3).

The rest of this document is organized as follows: the next section discusses previous studies over the same topics tackled in this work. Section~\ref{sec:dssGA} presents some background information about distributed GAs. Section~\ref{sec:proposed} proposes our fresh way to understand PGA performance. Sections~\ref{sec:pbm} and~\ref{sec:exp} detail our benchmark and the experimental settings used in this paper, respectively, while Sections~\ref{sec:time} and~\ref{sec:fit} analyze the results from different points of view: wall-clock time and numerical performance. Finally, we will end this work with a summary of the results (Section~\ref{sec:summ}), general conclusions, and future research in light of our results (Section~\ref{sec:conclu}).

\section{Related Works}

This section focuses on existing scientific articles analyzing the behavior of island-based PGAs and related evolutionary algorithms. For this, we have performed a systematic review looking for papers dealing with our topics during the last ten years. ``Speed-up,'' ``distributed evolutionary algorithms,'' and ``island model'' are some of the terms that have been searched on traditional literature platforms on this domain (IEEE Xplore, ACM Digital Library, SCOPUS, Google Scholar...). 

Most of the papers found focused on solving a specific problem instead of analyzing the running algorithm to understand it: scheduling~\cite{kurdi2016effective}, clustering~\cite{agusti2012new}, satellite design~\cite{hu2014optimization}, and routing~\cite{abbasi2020efficient}. The main objective of these studies was very different from the proposed in our work; however, they allow us to conclude on the good health of PGAs nowadays for many application fields. This also means that the kind of analysis we do in our paper can benefit many practitioners. In this sense, we could find several works analyzing the behavior from two different perspectives: more theoretical studies and more experimental analysis.

From a formal point of view, the authors in~\cite{lassig2014general} presented a method for  analyzing the run time of parallel evolutionary algorithms with a spatially structured population. This allowed an estimate of the speed-up under different migration topologies. Moreover, in~\cite{lassig2013design}, the same authors studied the influence of some migration parameters using a theoretical run time analysis. Both papers improved the body of knowledge about parallel metaheuristic based on the island model. However, they imposed some assumptions which usually are not found in real-world scenarios, limiting their impact. This is a reason to also go for mixed studies like this present one, where we combine actual thousands of experimental runs with a mathematical understanding of their shown behavior. In the rest of this section, we will focus on a more experimental analysis similar to the proposed one in this work.

We can distinguish two kinds of analysis for parallel metaheuristics: on the one hand, some papers analyzed the effect of the physical parallel platform (e.g., GPUs, multi-cores, cloud), while on the other hand, other works studied the design and the influence of the parameters in the performance. With respect to the first ones, we can cite~\cite{abdelhafez2019performance} for multiprocessors,~\cite{janssen2019acceleration} for GPUs,  and~\cite{khalloof2019superlinear} for cloud platforms. All of them showed the advantages of using parallel models for reducing the execution time when using a parallel platform. For example, in the last-mentioned paper~\cite{khalloof2019superlinear}, the authors showed how using microservices, container virtualization, and the publish/subscribe messaging paradigm could even lead to super-linear speed-up for a moderate number of islands (less than 64, after that the speed-up decreased). This present paper uses a different hardware platform: a multi-core computer cluster interconnected by high-speed network technology.

Finally, we end this section by describing some recent works which discussed the influence of the migration policy on the performance of this kind of parallel technique. This was a difficult choice since many migration parameters exist, but simultaneously, each one will multiply a lot of the needed experimental runs for the later mathematical analysis. One of the most recent works studying migration policies can be found in~\cite{wang2019empirical}. In this paper, the authors examined the solution quality, convergence speed, and population diversity of the island model EA with different migration topologies, population sizes, migration rates, and migration gaps on four benchmark functions of 1,000 dimensions.  Their results showed the relation between migration rate and migration gap, and the high importance of the migration rate and the topology. In our work, we examine the migration gap due to its influence on the results, however, we analyze a genetic algorithm in contrast to the differential evolution used in that work. Others papers like~\cite{muszynski2015distributed} perform similar studies, but they focus on other issues like fault tolerance.

\section{Island-based Distributed Steady-State GA~\label{sec:dssGA}}
This section explains the PGA technique used in this work. In particular, we use a standard distributed island-based steady-state GA (dssGA)~\cite{Tomassini2006}. We choose this type of algorithm since it is one of the most popular PGAs used in the literature and the Industry~\cite{harada2020parallel}, thus improving our knowledge on it could produce a high impact in multiple domains.

\begin{algorithm}[tb]
\centering
\caption{A pseudo-code of island-based distributed steady-state GA in the $i$-th island}
\label{alg:pseudo-code}
\begin{algorithmic}[1]
\State $t\leftarrow 0$
\State $\texttt{initialization}(P_0)$
\State $\texttt{evaluation}(P_0)$
\While{Termination criterion}
    \State $p_1, p_2 \leftarrow \texttt{selectParent}(P_t)$
    \If{$rand(0, 1)<P_c$} \Comment{Crossover probability}
        \State $p_1^\prime \leftarrow \texttt{crossover}(p_1, p_2)$
    \Else
        \State $p_1^\prime \leftarrow p_1$
    \EndIf
    \If{$rand(0, 1)<P_m$}\Comment{Mutation probability}
        \State $p_1^{\prime\prime} \leftarrow \texttt{mutation}(p_1^\prime)$
    \Else
        \State $p_1^{\prime\prime} \leftarrow p_1^\prime$
    \EndIf
    \State $\texttt{evaluation}(p_1^{\prime\prime})$
    \State $ind\leftarrow \texttt{randomSelection}(P_t)$
    \If{$f(p_1^{\prime\prime})<f(ind)$}\Comment{Minimization}
        \State $P_{t+1}\leftarrow P_t \cup \{p_1^{\prime\prime}\} \setminus \{ind\}$ \Comment{Replacement}
    \EndIf
    \If{$t \% \Delta T$} \Comment{Migration gap}
        \State $mig \leftarrow \texttt{migrantSelection}(P_{t+1})$ 
        \Statex \Comment{Select the best solution}
        \State Send $mig$ to $(i + 1)$-th island
    \EndIf
    \If{receive migrant}
        \Statex \Comment{Asynchronous communication}
        \State $mig \leftarrow \texttt{receive}()$
        \State $ind\leftarrow \texttt{randomSelection}(P_t)$
        \If{$f(mig)<f(ind)$}\Comment{Minimization}
            \State $P_{t+1}\leftarrow P_{t+1} \cup \{mig\} \setminus \{ind\}$
        \Statex\Comment{Replacement}
        \EndIf
    \EndIf
    \State $t\leftarrow t+1$
\EndWhile
\end{algorithmic}
\end{algorithm}

Algorithm~\ref{alg:pseudo-code} shows a pseudo-code of the distributed island-based steady-state GA (dssGA) implemented in this work.
A dssGA consists of many subpopulations (islands) that evolve governed by a local steady-state GA in parallel.
First, the initial population is generated, and the required fitness values are evaluated. Then, a parent selection method (e.g., binary tournament selection) selects two parents, and crossover and mutation operations create a new offspring. Finally, a newly generated offspring is evaluated and compared with a randomly selected individual in the population. If the offspring is better than the random member, the offspring replaces it in the population.

Along with these iterative procedures, the migration operation is performed to communicate with other islands at a fixed interval (i.e., migration gap, $\Delta T$). The migration operation selects an individual (called migrant) from the local population and sends a copy of it to another island. When receiving a migrant from other islands, it is merged into the population if its fitness is better than an individual randomly selected from the population.

\begin{figure}[tb]
    \centering
    \includegraphics[width=0.8\linewidth]{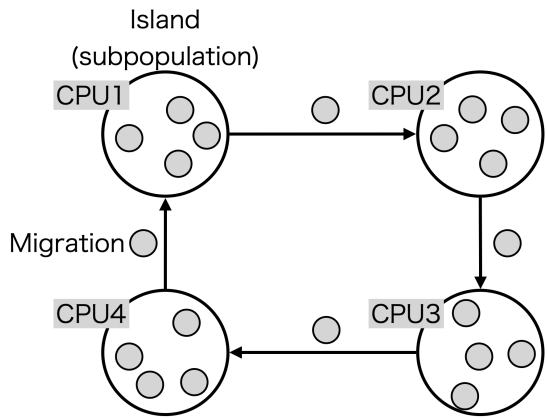}
    \caption{Illustration of the island-based GA}
    \label{fig:island_model}
\end{figure}

The migration gap $\Delta T$ (also known as the migration interval) determines how frequently each island communicates with the other~\cite{cantu1995summary}. The smaller the migration gap is used, the more frequent communication is performed.
The migration selection determines what individual is migrated to other islands. Although several migration selections have been explored (e.g., best, random)~\cite{cantu2001migration}, this paper considers the best solution selection that selects an individual having the best fitness in the island as a migrant.
The migration topology determines the connection between islands, and it highly impacts the search behavior of dssGA. Although several topologies have been proposed (e.g., uni- and bi-directional ring, random, hypercube)~\cite{Falco2014}, this paper considers the uni-directional migration topology, which is simple and the most popular one. Fig.~\ref{fig:island_model} shows an illustration of the uni-directional ring topology, where a migrant is sent to the island having the next index. The uni-directional ring topology is fixed during the optimization.

\section{Proposed Analysis Methodology}\label{sec:proposed}
This paper advances in how we can understand PGA performance beyond simple inspections of tables/graphs. In particular, since wall-clock time and speed-up are essential features in analyzing PGAs, it is worth going beyond previous works and mathematically model these features for running algorithms.
Usually, wall-clock time measures the running time of a PGA until a given  termination criterion is met, e.g., reaching a maximum number of fitness evaluations or finding a good-enough solution.
On the other hand, speed-up represents a wider point of view, allowing the understanding of the scalability of the algorithm and even its quality as a piece of software. For that, speed-up studies how wall-clock time evolves when using a varying number of $m$ processors. The speed-up $S(m)$ when using $m$ processors is calculated as:
\begin{eqnarray}
S(m)=\frac{\tilde{T}_1}{\tilde{T}_m}\label{eq:speed-up}
\end{eqnarray}
where $T_m$ represents the wall-clock time when the PGA is run on $m$ cores, ideally, the speed-up is upper bounded by the number of processing units, $S(m)=m$ (linear speed-up). However, in practice, it results in smaller values (sub-linear speed-up) because of the added communication times between cores and some parts of the algorithm that cannot be parallelized. Thus, sub-linear values above, e.g., 80\% of the ideal value, are usually accepted as good ones. On the contrary, an intriguing super-linear speed-up value has been shown to be also possible for PGAs ($S(m)>m$)~\cite{ALBA2001451} in some scenarios due to more efficient use of the hardware resources (e.g., when subpopulations now fit into caches and are no longer stored in main memory for access) and because the distributed search is experiencing new drifting effects or new paths leading till the optimum solution. 

Speed-up is usually studied as numeric figures in a table or as a plotted line in a graph. However, we would like to propose the routinary use of a mathematical understanding of the observed speed-up as a way of fostering a more explainable artificial intelligence. The models that we will use for this analysis are chosen as a result of the preliminary study (see our Appendix) and common mathematical knowledge after inspecting the resulting graphs in the literature~\cite{ROCHA2014288,LIU201598}. In particular, we compare several families of mathematical functions: linear, exponential, logarithmic, and three variants of rational equations.
From the preliminary study, the rational function of $f(x)=(ax+b)/(x+c)$ shows the lowest numerical error (and a reasonably low number of free parameters) for both predictions of the wall-clock time and the speed-up.
For this reason, this paper uses this rational function for understanding the PGA performance. Please see the detail of the model selection presented in Appendix~\ref{sec:app_fun}.

We will then first study the wall-clock time of a PGA by analyzing its relationship with the number of cores as the rational equation of:
\begin{equation}
    \hat{T}(x)=\frac{ax+b}{x+c},\label{eq:wct_model}
\end{equation}
where $x$ indicates the number of used cores, and $\hat{T}(x)$ is the fitted (predicted) model for the wall-clock time; $a, b$, and $c$ are the model-free parameters.
According to this model, $a$ indicates the limit value of the wall-clock time when the number of used cores approaches infinity, i.e., $x\rightarrow\infty$.
If $a=0$ and $c=0$ (our experiment will show that these values tend to be quite small), this model is denoted as $\hat{T}(x)=b/x$. This indicates an ideal behavior, where the wall-clock time when using one core is $b$, and it decreases as the number of cores increases. In Eq.~\eqref{eq:wct_model}, the predicted wall-clock time converges $\lim_{x\rightarrow\infty}\hat{T}(x)=a$ where the number of used cores approaches infinity. This analysis will be provided in Section~\ref{sec:understand_wct}.

On the other hand, as to the speed-up performance, we propose a model of the rational function as:
\begin{equation}
    \hat{S}(x)=\frac{ax+b}{x+c},\label{eq:rat_model}
\end{equation}
where $x$ indicates the number of used cores, and $\hat{S}(x)$ indicates the predicted speed-up performance; $a, b,$ and $c$ are the model-free parameters.
This model is similar to the model used in the wall-clock time. In particular, when the wall-clock time is modeled as $\hat{T}(x)=(ax+b)/(x+c)$ as Eq.~\eqref{eq:wct_model}, we can calculate the speed-up as $\hat{S}(x)=\hat{T}(1)/\hat{T}(x)\equiv (a^\prime x+b^\prime)/(x+c^\prime)$ from the definition of speed-up in Eq.~\eqref{eq:speed-up}.
In Eq.~\eqref{eq:rat_model}, the limit value of the predicted speed-up converges to $\lim_{x\rightarrow\infty}\hat{S}(x)=a$ where the number of used cores approaches infinity.
This analysis will be provided in Section~\ref{sec:understand_su}.

The benefits of offering a mathematical model of these important PGA values are manifold: (i) they could allow predicting the performance for the unknown number of processor units without testing on them, (ii) they allow analyzing the speed-up and wall-clock time by using only a few parameters (three, in our model) instead of multiple points/values in a curve/table, and (iii) they allow making easier/quantitative comparisons between algorithms. To summarize, a mathematical model for these metrics provides a powerful tool to understand and compare PGAs.

\section{Benchmark Problems\label{sec:pbm}}
This paper uses two well-known benchmark problems: the P-PEAKS problem~\cite{PPEAKS} and the vehicle routing problem (VRP)~\cite{VRP}. Both of them are discrete combinatorial problems but represent very different search features found in many other combinatorial tasks. First, regarding the problem class, both problems are NP-Hard~\cite{garey1979computers}, thus usually requiring metaheuristic algorithms. Second, P-PEAKS is encoded as a binary vector, while VRP needs a different permutation-based encoding. These problems then represent two huge classes of combinatorial problems solved with PGAs. Finally, if we focus on the computational complexity of their fitness functions, P-PEAKS is quadratic ($O(n^2)$) while VRP is linear ($O(n)$); whether this is relevant for the time and speed-up of the algorithms remains to be analyzed. Again, a large number of problems have similar complexities in their fitness functions, thus completing a good handful of reasons to use them for evaluating PGAs (our primary goal, we are not solving these problems as the main goal in this study). The following subsections show brief descriptions of these problems. 

\subsection{P-PEAKS}
The P-PEAKS problem~\cite{PPEAKS} is a binary optimization problem. P-PEAKS has a set of $P$ random $N$-bit strings, representing the location of $P$ (abstract) peaks in the search space. The fitness of the bit string is calculated as:
\begin{equation}
f(x)=\min_{i=1, \cdots, P}\textrm{Hamming}(x, Peak_i),
\end{equation}
where $\textrm{Hamming}(x, Peak_i)$ calculates the Hamming distance between the bit string $x$ and the $i$-th peak $Peak_i$. The optimum fitness is 0.
We define a small problem by setting $P=20$ and $N=100$, while a large problem is set as $P=200$ and $N=1000$. The termination criteria for P-PEAKS problems are:
\begin{itemize}
\item P-PEAKS 20-100
\begin{itemize}
\item Termination criterion: Find optimum solution. $fitness=0$
\end{itemize}
\item P-PEAKS 200-1000
\begin{itemize}
\item Termination criterion: Find optimum solution $fitness=0$
\end{itemize}
\end{itemize}

\subsection{VRP}
The vehicle routing problem~\cite{VRP} (VRP) consists of determining minimum cost routes for a fleet of vehicles originating and terminating in a depot. The fleet of vehicles gives service to a set of customers with a known set of constraints. All customers must be assigned to vehicles such that each customer is serviced exactly once, taking into account the limited capacity of each vehicle. VRP encodes in each permutation the order in which customers are served. We use two different sizes (i.e., the number of customers) of VRPs from \cite{Christofides1979}. The number of customers, the best-known cost, and the termination criteria we use are:
\begin{itemize}
    \item VRP1
    \begin{itemize}
        \item \# of customers: 50
        \item Best known cost $fitness=524.61$
        \item Termination criterion: Find a solution with $\allowbreak fitness=650$
    \end{itemize}
    \item VRP2
    \begin{itemize}
        \item \# of customers: 75
        \item Best known cost $fitness=835.26$
        \item Termination criterion: Find a solution with $\allowbreak fitness=900$
    \end{itemize}
\end{itemize}

\section{Experimental Settings\label{sec:exp}}
This section presents experimental settings, including the system specifications and used parameters. First, we discuss the system hardware and software. Next, we describe the parameter settings regulating the island model dssGA algorithm. 

\subsection{System Specifications}
\begin{table}[!tb]
    \centering
    \caption{Utilized computer specifications}
    \label{tab:spec}
    \begin{tabular}{l|ll}
    \hline
    PC1 &CPU&Intel(R) Xeon(R) E5-1650 v2\\
    &&(3.50GHz, 6 cores)\\
    &Memory&16GB  \\
    &OS&Ubuntu 18.04.5 LTS\\
    &Network&1Gbps\\
    \hline
    PC2 &CPU&Intel(R) Xeon(R) E5-1650 v2\\
    &&(3.50GHz, 6 cores)\\
    &Memory&16GB  \\
    &OS&Ubuntu 18.04.5 LTS\\
    &Network&1Gbps\\
    \hline
    PC3 &CPU&Intel(R) Xeon(R) E5-1650 v2\\
    &&(3.50GHz, 6 cores)\\
    &Memory&16GB  \\
    &OS&Ubuntu 18.04.5 LTS\\
    &Network&1Gbps\\
    \hline
    PC4 &CPU&Intel(R) Xeon(R) E5-1650 v3\\
    &&(3.50GHz, 6 cores)\\
    &Memory&16GB  \\
    &OS&Ubuntu 18.04.5 LTS\\
    &Network&1Gbps\\
    \hline
    PC5 &CPU&Intel(R) Xeon(R) E5-1650\\
    &&(3.20GHz, 6 cores)\\
    &Memory&16GB  \\
    &OS&Ubuntu 18.04.5 LTS\\
    &Network&1Gbps\\
    \hline
    PC6 &CPU&Intel(R) Xeon(R) E5-1650\\
    &&(3.20GHz, 6 cores)\\
    &Memory&16GB  \\
    &OS&Ubuntu 18.04.5 LTS\\
    &Network&1Gbps\\
    \hline
    \end{tabular}
\end{table}

Our experiments are executed on a cluster computing system consisting of six multi-core computers, as shown in Table~\ref{tab:spec}.
We use Intel Xeon E5-1650 processors, 16 GB of RAM, Linux Ubuntu 18.04.5 LTS operating system, connected through a 1 Gbps Ethernet.
We compare a wealth of dssGA algorithms over the different number of cores, in particular, 1, 2, 4, 8, 16, 32, and 64 cores are used, and their behavior is compared and mathematically modeled.
In order to utilize six computers uniformly, our experiments configure the number of utilized cores as shown in Table~\ref{tab:utilized_cores}.
Our algorithms are implemented in C++/MPI. We choose Open MPI~\cite{openmpi}, an open-source MPI implementation that is very popular. 

\begin{table}[!tb]
    \centering
    \caption{The number of utilized cores}
    \label{tab:utilized_cores}
    \scalebox{0.9}{
    \begin{tabular}{l|llllll}
    \hline
    \# of utilized cores&PC1&PC2&PC3&PC4&PC5&PC6\\
    \hline
    1	&1	&0	&0	&0	&0	&0\\
    2	&1	&1	&0	&0	&0	&0\\
    4	&1	&1	&1	&1	&0	&0\\
    8	&2	&2	&1	&1	&1	&1\\
    16	&3	&3	&3	&3	&2	&2\\
    32	&6	&6	&5	&5	&5	&5\\
    64	&11	&11	&11	&11	&10	&10\\
    \hline
    \end{tabular}
    }
\end{table}

\subsection{Parameter Settings}
\begin{table}[!t]
    \centering
    \caption{Parameters used in the experiment}
    \label{tab:parameters_exp1}
    \begin{tabularx}{\columnwidth}{l|X|X}
    \hline
    \multicolumn{1}{c|}{Parameter}   &\multicolumn{2}{c}{Value}\\
    \cline{2-3}
    &\multicolumn{1}{c|}{P-PEAKS}&\multicolumn{1}{c}{VRP}\\
    \hline
    \# of islands     &  \multicolumn{2}{l}{64}\\
    Population size     & \multicolumn{2}{l}{16/island}\\
    Selection   & \multicolumn{2}{l}{Binary tournament selection}\\
    Migration gap    & \multicolumn{2}{l}{\{16, 32, 64, 128, 256\} evals.}\\
    Migration selection    & \multicolumn{2}{l}{Best solution}\\
    Migration topology  & \multicolumn{2}{l}{Uni-directional ring}\\
    \hline
    Crossover type  & Single point crossover&Cycle crossover\\
    Crossover probability   & 0.8&0.8\\
    Mutation type    & Bit flip&Exchange\\
    Mutation probability    & 0.1&1.0\\
    \hline
    \end{tabularx}
\end{table}

In the experiments, we will analyze the effect of the migration gap for values 16, 32, 64, 128, and 256 (evaluations) in dssGA. 
The rest of the parameter setting used is shown in Table~\ref{tab:parameters_exp1}.

The number of islands is 64 (constant) regardless of the number of used cores. The population size is 16 for each island. A binary tournament selection is performed to pick the parents. These parameters are common both for P-PEAKS and VRP.
Then particular operators for crossover and mutation are needed for each problem due to their different encoding. For the P-PEAKS problem, the single-point crossover and the bit flip mutation are applied to generate a new offspring, with the probabilities $P_c=0.8$ and $P_m=0.1$, respectively. For the VRP, the cycle crossover~\cite{oliver1987study} and the exchange mutation~\cite{banzhaf1990} are applied, with probabilities $P_c=0.8$ and $P_m=1.0$, respectively.

\subsection{Evaluation Criteria\label{sec:eval_crit}}
For each experiment in this work, we make 30 independent runs and then evaluate the resulting numerical and computational effort. In particular, we will present results on the wall-clock time and the number of fitness evaluations until satisfying the termination condition.
In addition to these basic performance analyses, we study the speed-up, and approach all these outputs with numeric tables, trend figures, and mathematical formulae matching them.
This paper calculates the speed-up ($S$) as shown in Eq.~\eqref{eq:speed-up} using the median wall-clock time. Speed-up is an important metric since it gives an idea about the gains obtained when new computational processors are added to the system used by the algorithm and suggests what would happen in the future. It also allows judging the quality of the parallelization done on the algorithm, pointing out asymptotic behaviors.

We will calculate the median value and its standard deviation in 30 trials.
In each setting, the same random seeds are used to ensure the fairness of each experiment.
In addition, for all evaluation criteria, we perform the Kruskal-Wallis test~\cite{Kruskal1952} to confirm the statistical difference between the different migration gaps across the varying number of cores.

\subsection{Initial Data on Fitness Evaluation}
As a way of embarking on this study with basic information in our hands, we have first measured the average evaluation time for each problem fitness function. For this analysis, we have run $10^6$ solution evaluations on PC1 with the specification shown in Table~\ref{tab:spec}. Average evaluation times are shown in Table~\ref{tab:ave_eval_time}.
\begin{table}[!tb]
\centering
\caption{Average evaluation time for each problem [\si{\micro}s]}
\label{tab:ave_eval_time}
\begin{tabular}{lr}
\hline
Problem&Ave. Time [\si{\micro}s]\\
\hline
P-PEAKS 20-100&30.90\\
P-PEAKS 200-1000&3060.97\\
\hline
VRP1&2.69\\
VRP2&4.05\\
\hline
\end{tabular}
\end{table}

As to the results of this basic first study, the average evaluation time for P-PEAKS is 30.90 \si{\micro}s for the small problem and 3060.97 \si{\micro}s for the large one. The large problem is then $100$ times slower in evaluating solutions, although it is only ten times larger. This behavior was expected since the computational complexity of its fitness function ($O(n^2)$). In addition to shedding some light on the relationship between problem running times with actual data, it allowed us to confirm that our implementation runs as expected, and that theory matches practice. Even if simple, these conclusions are hardly found in previous works on PGAs.

\begin{table*}[!bt]
    \caption{Wall-clock time [s] (P-PEAKS). Boldface: minimum in column. Underline: minimum in row.}
    \label{tab:wct_ppeaks}
    \centering
    \scalebox{0.9}{
    \begin{tabular}{rl||rrrrrrrr!{\bvline{1pt}}rrrrrrrr}
    \hline
    &&\multicolumn{8}{c!{\bvline{1pt}}}{P-PEAKS 20-100}&\multicolumn{8}{c}{P-PEAKS 200-1000 $(\times10^3)$}\\
      Mig.   &&\multicolumn{7}{c}{\# of utilized cores}&&\multicolumn{7}{c}{\# of utilized cores}&\\
        \cline{3-18}
     gap    &&  1&2&4&8&16&32&64&SD&  1&2&4&8&16&32&64&SD\\
    \hline
    16&median&{{6.22}}&{{3.07}}&{{1.53}}&{{0.80}}&{{0.42}}&{{0.21}}&\underline{{0.13}}&**&{\textbf{6.25}}&{\textbf{3.21}}&{{1.69}}&{\textbf{0.88}}&{{0.46}}&{{0.23}}&\underline{{0.14}}&**\\
    &stdv.&0.25&0.16&0.07&0.03&0.02&0.01&0.01&&0.08&0.14&0.04&0.02&0.01&0.01&0.01&\\
    \hline
    32&median&{{6.29}}&{{3.03}}&{{1.51}}&{{0.79}}&{{0.41}}&{\textbf{0.20}}&\underline{{0.12}}&**&{{6.35}}&{{3.29}}&{\textbf{1.67}}&{{0.89}}&{\textbf{0.45}}&{{0.23}}&\underline{{0.14}}&**\\
    &stdv.&0.26&0.15&0.10&0.03&0.02&0.01&0.01&&0.11&0.08&0.04&0.02&0.01&0.01&0.01&\\
    \hline
    64&median&{{6.16}}&{{2.99}}&{{1.55}}&{{0.79}}&{{0.41}}&{{0.20}}&\underline{{0.12}}&**&{{6.47}}&{{3.39}}&{{1.68}}&{{0.88}}&{{0.46}}&{\textbf{0.23}}&\underline{{0.14}}&**\\
    &stdv.&0.27&0.17&0.07&0.04&0.02&0.01&0.01&&0.11&0.09&0.04&0.02&0.01&0.01&0.01&\\
    \hline
    128&median&{{6.19}}&{{3.05}}&{{1.53}}&{{0.79}}&{{0.41}}&{{0.21}}&\underline{{0.12}}&**&{{6.81}}&{{3.47}}&{{1.70}}&{{0.89}}&{{0.46}}&{{0.23}}&\underline{{0.14}}&**\\
    &stdv.&0.29&0.15&0.08&0.04&0.02&0.01&0.01&&0.15&0.08&0.06&0.02&0.01&0.01&0.00&\\
    \hline
    256&median&{{6.17}}&{{3.10}}&{{1.55}}&{{0.78}}&{{0.42}}&{{0.21}}&\underline{{0.12}}&**&{{7.17}}&{{3.63}}&{{1.75}}&{{0.91}}&{{0.47}}&{{0.24}}&\underline{{0.14}}&**\\
    &stdv.&0.30&0.14&0.06&0.04&0.02&0.01&0.01&&0.20&0.09&0.04&0.02&0.01&0.01&0.01&\\
    \hline
    \hline
    &SD&NS&NS&NS&NS&NS&**&NS&&**&**&**&**&**&**&NS&\\
    \hline
    \multicolumn{10}{l}{NS: No significant difference, *: $p<0.05$, **: $p<0.01$}
    \end{tabular}
    }
\end{table*}
\begin{table*}[tb]
    \centering
    \caption{Wall-clock time [s] (VRP). Boldface: minimum in column. Underline: minimum in row.}
    \label{tab:wct_vrp}
    \scalebox{0.9}{
    \begin{tabular}{rl||rrrrrrrr!{\bvline{1pt}}rrrrrrrr}
    \hline
    &&\multicolumn{8}{c!{\bvline{1pt}}}{VRP1}&\multicolumn{8}{c}{VRP2}\\
      Mig.   &&\multicolumn{7}{c}{\# of utilized cores}&&\multicolumn{7}{c}{\# of utilized cores}\\
        \cline{3-18}
     gap    &&  1&2&4&8&16&32&64&SD&  1&2&4&8&16&32&64&SD\\
    \hline
    16&median&{{3.95}}&{\textbf{2.06}}&{{1.07}}&{\textbf{0.57}}&{{0.30}}&{{0.18}}&\underline{{0.15}}&**&{{9.90}}&{{5.34}}&{\textbf{2.67}}&{{1.50}}&{\textbf{0.76}}&{\textbf{0.44}}&\underline{{0.38}}&**\\
    &stdv.&0.52&0.33&0.19&0.08&0.04&0.03&0.04&&0.86&0.69&0.67&0.34&0.13&0.05&0.07&\\
    \hline
    32&median&{\textbf{3.85}}&{{2.14}}&{{1.12}}&{{0.59}}&{{0.30}}&{{0.18}}&\underline{{0.15}}&**&{\textbf{9.75}}&{\textbf{5.31}}&{{2.82}}&{\textbf{1.47}}&{{0.80}}&{{0.44}}&\underline{{0.38}}&**\\
    &stdv.&0.44&0.26&0.18&0.08&0.05&0.03&0.03&&1.56&0.73&0.46&0.18&0.10&0.05&0.06&\\
    \hline
    64&median&{{3.97}}&{{2.07}}&{\textbf{1.03}}&{{0.58}}&{\textbf{0.30}}&{{0.19}}&\underline{{0.15}}&**&{{10.64}}&{{5.59}}&{{2.87}}&{{1.57}}&{{0.82}}&{{0.48}}&\underline{{0.39}}&**\\
    &stdv.&0.32&0.32&0.16&0.12&0.04&0.03&0.03&&1.38&0.60&0.39&0.17&0.09&0.07&0.06&\\
    \hline
    128&median&{{4.10}}&{{2.23}}&{{1.15}}&{{0.63}}&{{0.33}}&{{0.19}}&\underline{{0.16}}&**&{{10.87}}&{{5.81}}&{{2.96}}&{{1.60}}&{{0.83}}&{{0.47}}&\underline{{0.39}}&**\\
    &stdv.&0.51&0.32&0.18&0.12&0.05&0.03&0.03&&1.47&0.82&0.46&0.23&0.10&0.05&0.06&\\
    \hline
    256&median&{{4.59}}&{{2.30}}&{{1.18}}&{{0.64}}&{{0.36}}&{{0.20}}&\underline{{0.16}}&**&{{12.06}}&{{6.28}}&{{3.22}}&{{1.75}}&{{0.91}}&{{0.48}}&\underline{{0.40}}&**\\
    &stdv.&0.48&0.28&0.11&0.10&0.04&0.03&0.03&&1.11&0.81&0.43&0.17&0.11&0.07&0.05&\\
    \hline
    \hline
    &SD&**&*&*&**&**&NS&NS&&**&**&**&**&**&*&NS&\\
    \hline
    \multicolumn{10}{l}{NS: No significant difference, *: $p<0.05$, **: $p<0.01$}\\
    \end{tabular}}
\end{table*}
All VRPs take a shorter evaluation time than the smaller P-PEAKS problem. The evaluation time of VRP1 is 2.69 \si{\micro}s, while that of VRP2 is 4.05 \si{\micro}s.
VRP2 takes less than twice a longer evaluation time than VRP1. Again, these values can be explained by the linear complexity of its fitness function.

Equipped with this initial knowledge on the fitness function evaluation of the PGAs, we now face an explanation of the computational effort.

\section{Analysis of the Computational Effort\label{sec:time}}
In this section, we present several analyses from the viewpoint of the computational effort. This result shows the run time cost of dssGA with different migration gaps and the different number of cores. First, we measure the wall-clock time for different problems, migration gaps, and the different number of used cores. Then, we present the relation between the number of used cores and the wall-clock time and discuss the curve fitting with the rational function shown in Eq.~\eqref{eq:wct_model}.

\subsection{Measuring Wall-clock Time}
This subsection spins around the wall-clock time needed until satisfying the termination criteria shown in Section~\ref{sec:pbm}. 
Tables~\ref{tab:wct_ppeaks} and \ref{tab:wct_vrp} show the wall-clock time of P-PEAKS 20-100, P-PEAKS 200-1000, VRP1, and VRP2, respectively. The ``median'' row indicates the median wall-clock time, while the ``stdv.'' row indicates the standard deviation. The results of the statistical tests are shown in the ``SD'' row and column. The smallest median values in each column (between the different migration gaps) are shown in boldface if a significant difference is found. Meanwhile, the smallest values in each row (across all number of used cores) are denoted with an underlined text.

From the results got on P-PEAKS 20-100, we cannot extract significant differences between the different migration gaps, except when 32 cores are used. When using 32 cores, the migration gap of 32 shows the shortest wall-clock time. 
On P-PEAKS 200-1000, a significant difference between the migration gaps is found when using 32 or fewer cores, while no significant difference is found when using 64 cores.
When using 32 or fewer cores, the smaller migration gaps, i.e., the migration gap of 16 and 32, significantly achieve a shorter wall-clock time than the others.

A comparison between the different sizes of P-PEAKS problems thus reveals that a larger size of the P-PEAKS problem is highly affected by the migration gap. This indicates that the migration gap has a high impact on more complex problems.

The result on VRP1 shows that a significant difference between the different migration gaps is found when using 16 or fewer cores, while no significant difference is found when using more.
On VRP1, the smaller migration gaps also show better performance. In particular, the migration gaps of 16, 32, and 64 all achieve the shortest wall-clock time. 
The result on VRP2 shows that a significant difference between the different migration gaps is found when using 32 or fewer cores, while no significant difference is found when using 64 cores. On VRP2, the smaller migration gaps also show better performance. In particular, the migration gaps of 16 and 32 show the shortest wall-clock time.

By comparing the differently sized VRPs, a tendency can be found. The smaller migration gap obtains the shorter wall-clock time. In addition, VRP2 takes twice a longer wall-clock time than VRP1, which means that the difference in the evaluation time shown in Table~\ref{tab:ave_eval_time} is directly reflected later in the total wall-clock time. That is an interesting finding that could open new research lines to find out how much the fitness function affects the global times in practice for related classes of problems.

These results show that a small migration gap (i.e., 64 or less) tends to reduce the wall-clock time for all problems. On the other hand, when using a large number of cores (i.e., 64 cores), the effect of the migration gap on the wall-clock time is not significant.

A significant difference is found in all migration gaps and all problems as to the different number of utilized cores. In all problems, the wall-clock time is reduced by using a large number of cores, as expected in a well-programmed PGA. In particular, the wall-clock time when using 64 cores achieves the shortest wall-clock time. Even if it seems straightforward, implementing a PGA involves many programming decisions that can easily ruin the resulting running time.

To summarize, we have observed that, as we expected, the number of cores is the main factor in reducing the execution time. However, when the available hardware resources are reduced (a lower number of cores) and/or the problem is more complex, the utilization of a more coupled island model (a lower migration gap) also helps to reduce the wall-clock time. This is because a lower migration gap fosters more interchange of information between the islands, and faster convergence. We will verify this in Section~\ref{sec:fit}, which analyzes the number of fitness evaluations needed to find a high-quality solution.

\subsection{Understanding Wall-clock Time}\label{sec:understand_wct}
In this subsection, we show the boxplot of the wall-clock time and provide a mathematical model for predicting its trend. As the mathematical model, we use the rational function shown in Eq.~\eqref{eq:wct_model} and perform a curve fitting with the median wall-clock times. 

\begin{figure*}[tb]
\begin{tabular}{cc}
\begin{minipage}[t]{0.49\textwidth}
    \centering
    \includegraphics[scale=0.6]{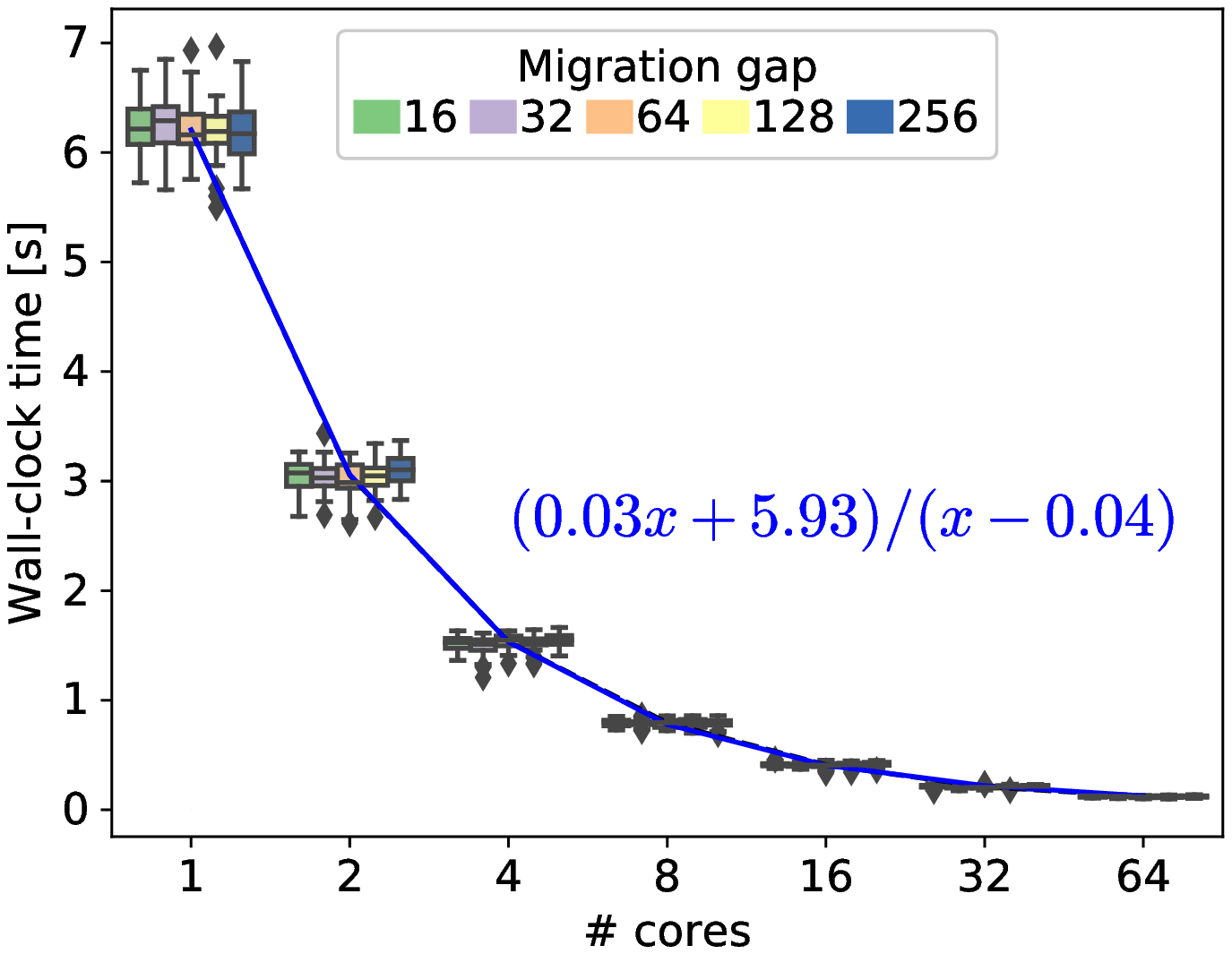}
    \caption{The boxplot of the wall-clock time on P-PEAKS 20-100 [s]}
    \label{fig:time_med_exp1}
\end{minipage}&
\begin{minipage}[t]{0.49\textwidth}
    \centering
    \includegraphics[scale=0.6]{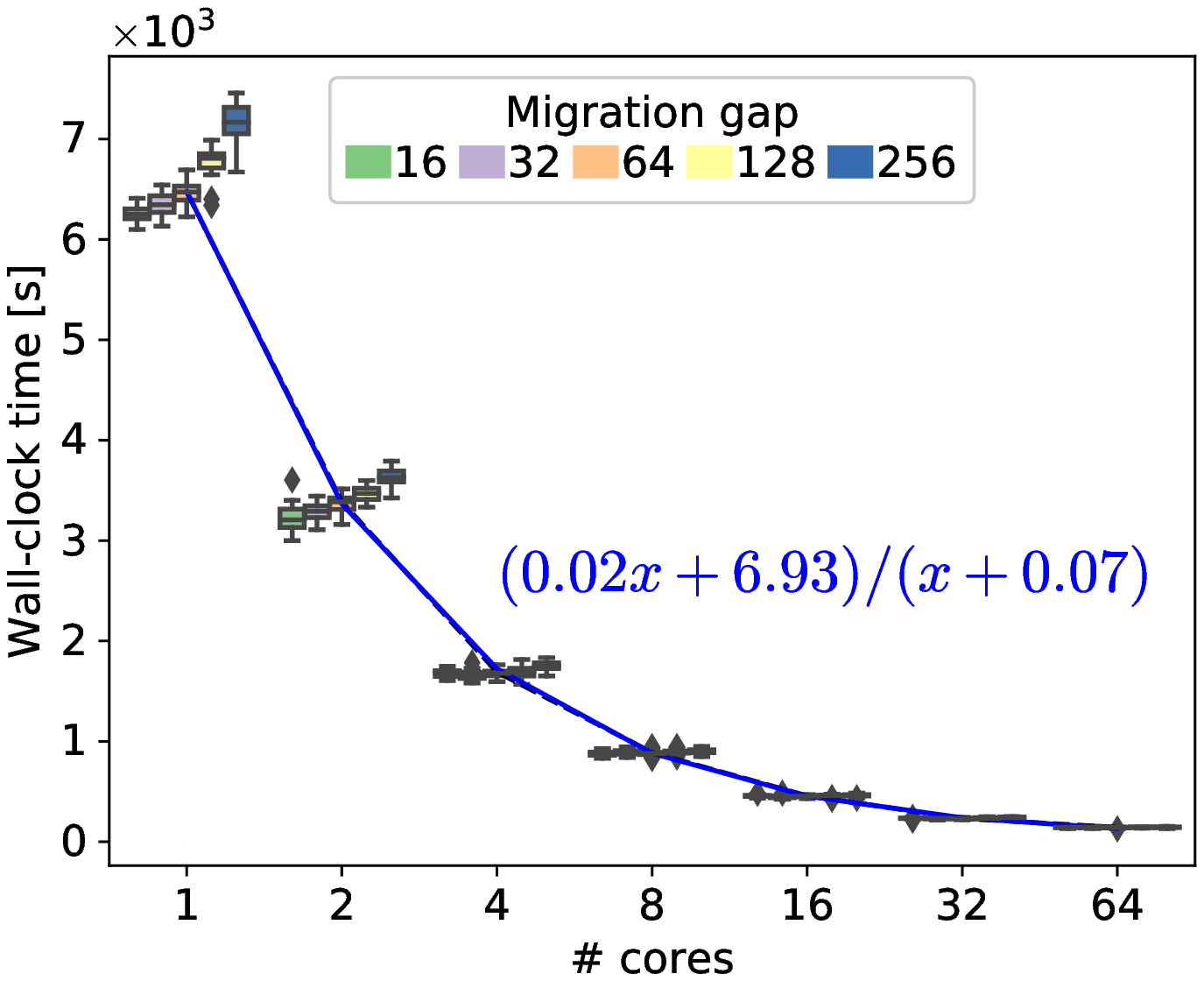}
    \caption{The boxplot of the wall-clock time on P-PEAKS 200-1000 [s]}
    \label{fig:time_med_exp1_large}
\end{minipage}\\
\begin{minipage}[t]{0.49\textwidth}
    \centering
    \includegraphics[scale=0.6]{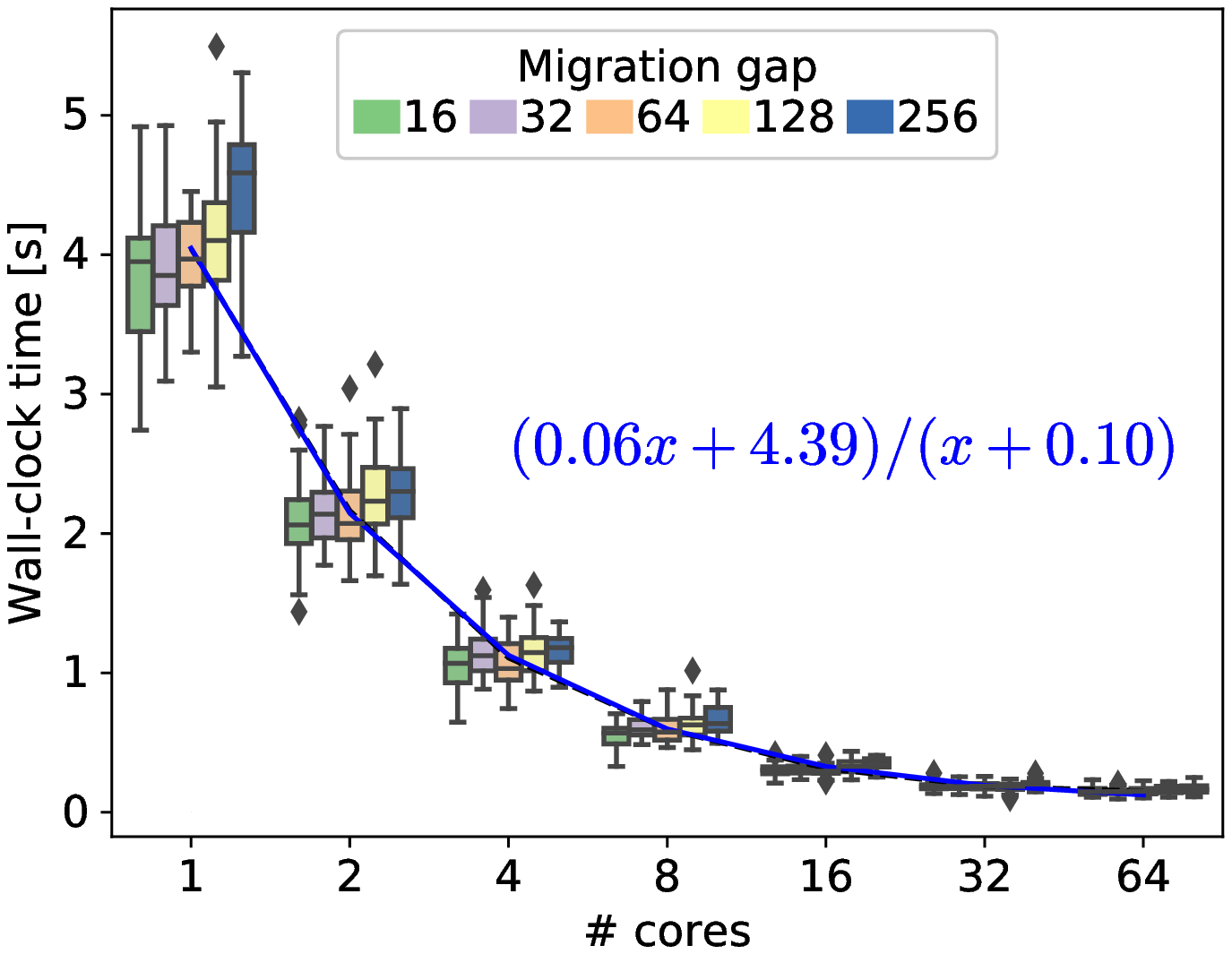}
    \caption{The boxplot of the wall-clock time on VRP1 [s]}
    \label{fig:time_med_exp1_vrp1}
\end{minipage}&
\begin{minipage}[t]{0.49\textwidth}
    \centering
    \includegraphics[scale=0.6]{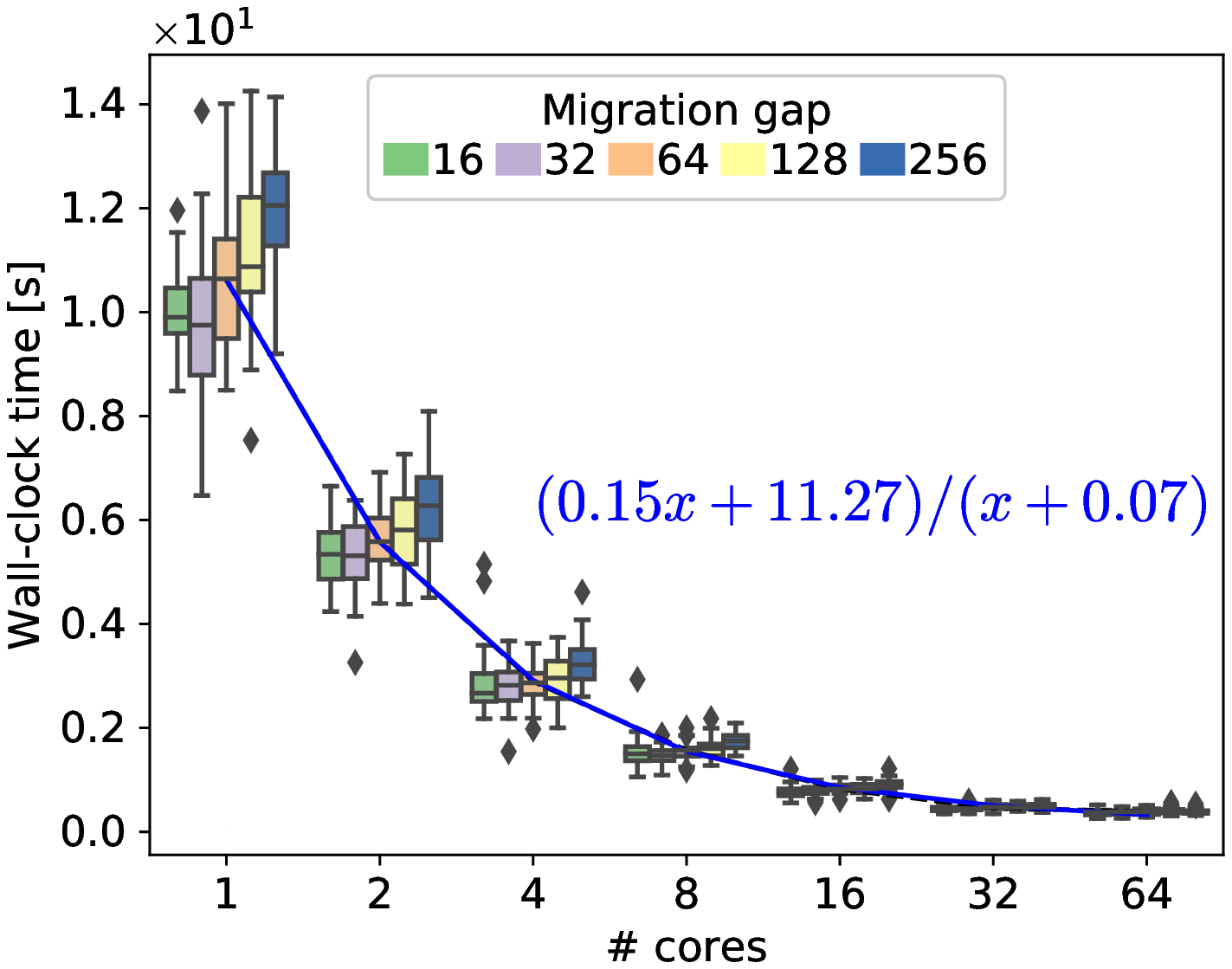}
    \caption{The boxplot of the wall-clock time on VRP2 [s]}
    \label{fig:time_med_exp1_vrp2}
\end{minipage}
\end{tabular}
\end{figure*}
Figs.~\ref{fig:time_med_exp1}--\ref{fig:time_med_exp1_vrp2} show the boxplot of the wall-clock time on P-PEAKS 20-100, P-PEAKS 200-1000, VRP1, and VRP2, respectively.
The horizontal axis indicates the number of utilized cores, while the vertical axis indicates the wall-clock time.
The different colors show the difference in migration gaps. The solid black line connects the median of the wall-clock times for each number of cores, while the solid blue line indicates the fitted curve, as with Eq.~\eqref{eq:wct_model} (these two lines almost overlap).

From these figures, we can see that the wall-clock time decreases exponentially as the number of cores increases for all benchmark problems. 
For more detail, the result of P-PEAKS 20-100 in Fig.~\ref{fig:time_med_exp1} indicates that the difference in wall-clock time between different migration gaps is slight.
On the other hand, for P-PEAKS 200-1000 shown in Fig.~\ref{fig:time_med_exp1_large}, there is a positive correlation between the migration gaps, and the smaller the migration gap is used, the shorter wall-clock time is achieved.

For VRP1 and VRP2, we can find a similar tendency to P-PEAKS 200-1000. That is, the smaller migration gap achieves the shorter wall-clock time. However, compared with the P-PEAKS problems, the variance of the wall-clock time is larger on VRPs. Furthermore, it can be seen that the standard deviation when using 64 cores is the same, or larger in some cases, compared with that when using 32 cores. This is because the solution evaluation time is relatively fast (about ten times faster than P-PEAKS 20-100) than the algorithm execution, and the overhead of the operating system and network communication is significant against the execution time. Hence, it should be noted that for problems with short evaluation times, the adverse effects of increased computational overhead may dominate the benefits of parallelization.

The result of model fitting shows that the rational function well represents the change of the wall-clock time against the number of used cores. This model first gives us a good prediction of the limit value for the wall-clock time. In particular, the predicted wall-clock time converges to $a$ of the model when the number of used cores approaches infinity, i.e., $x\rightarrow \infty$. So, the predicted limit wall-clock times for P-PEAKS 20-100, P-PEAKS 200-1000, VRP1, and VRP2 are 0.03s, 20 ($=0.02\times 10^3$)s, 0.06s, and 0.15s, respectively. The parameter $c$ is small in all the experiments ($\pm 0.1$), and its influence is negligible when the number of cores ($x$) is larger than $2$. Finally, parameter $b$  mainly depends on the wall-clock time when the technique is executed in one core, and its influence is reduced when the number of cores increases. 

In summary, we can declare the $a$ parameter as the most representative value when the number of cores is medium or large, while $b$ is more important for a small number of cores. Parameter $c$ is almost negligible in all the cases, but we maintain it for the sake of homogeneity with our coming speed-up analysis.

\section{Analysis of the Numerical Effort\label{sec:fit}}
The last section was concerned with time, a feature of the computer program running the algorithm. In this section, we go for the number of fitness evaluations needed to solve a problem, a feature of the algorithm itself. This result shows the numerical effort of dssGA with different migration gaps over a growing number of cores.

\subsection{Measuring the Number of Fitness Evaluations}
This subsection discusses the number of fitness evaluations until satisfying the termination criteria shown in Section~\ref{sec:pbm}. 
\begin{table*}[tb]
    \centering
    \caption{The number of fitness evaluations (P-PEAKS). Boldface: minimum in column. Underline: minimum in row}
    \label{tab:eval_ppeaks}
    \scalebox{0.9}{
    \begin{tabular}{rl||rrrrrrrr!{\bvline{1pt}}rrrrrrrr}
    \hline
    &&\multicolumn{8}{c!{\bvline{1pt}}}{P-PEAKS 20-100 ($\times 10^3$)}&\multicolumn{8}{c}{P-PEAKS 200-1000 ($\times 10^4$)}\\
      Mig.   &&\multicolumn{7}{c}{\# of utilized cores}&&\multicolumn{7}{c}{\# of utilized cores}\\
        \cline{3-18}
     gap    & &  1&2&4&8&16&32&64&SD&  1&2&4&8&16&32&64&SD\\
    \hline
    16&median&{\textbf{185}}&{\textbf{186}}&{\textbf{183}}&{\textbf{186}}&{\textbf{185}}&{{177}}&\underline{{153}}&**&{\textbf{193}}&\underline{{191}}&{{213}}&{{216}}&{{221}}&{{213}}&{{201}}&**\\
    &stdv.&7.60&9.77&9.00&7.24&8.50&11.6&9.56&&2.45&30.6&4.73&5.66&5.23&5.15&7.77&\\
    \hline
    32&median&{{193}}&{{190}}&{{188}}&{{189}}&{{188}}&{\textbf{176}}&\underline{{154}}&**&{{202}}&\underline{\textbf{139}}&{\textbf{179}}&{\textbf{187}}&{\textbf{194}}&{\textbf{190}}&{\textbf{180}}&**\\
    &stdv.&7.78&9.14&12.3&8.08&8.27&11.0&10.5&&3.51&3.57&4.99&4.88&6.13&6.02&7.37&\\
    \hline
    64&median&{{192}}&{{189}}&{{195}}&{{192}}&{{191}}&{{179}}&\underline{{149}}&**&{{209}}&\underline{{145}}&{{182}}&{{187}}&{{200}}&{{191}}&{{183}}&**\\
    &stdv.&8.19&10.8&8.84&10.3&11.7&10.1&11.4&&3.63&3.74&4.22&5.34&4.28&4.97&6.44&\\
    \hline
    128&median&{{194}}&{{195}}&{{194}}&{{192}}&{{196}}&{{188}}&\underline{{154}}&**&{{222}}&\underline{{151}}&{{186}}&{{191}}&{{204}}&{{196}}&{{186}}&**\\
    &stdv.&9.13&9.73&10.0&11.1&9.54&10.7&12.2&&4.80&14.4&6.30&4.60&4.95&6.49&5.50&\\
    \hline
    256&median&{{194}}&{{199}}&{{196}}&{{191}}&{{198}}&{{190}}&\underline{{156}}&**&{{234}}&\underline{{161}}&{{191}}&{{197}}&{{206}}&{{200}}&{{189}}&**\\
    &stdv.&9.22&9.07&8.40&10.1&10.8&9.76&10.7&&6.43&30.6&4.72&7.20&5.79&7.69&6.58&\\
    \hline
    \hline
    &SD&**&**&**&*&**&**&NS&&**&**&**&**&**&**&**&\\
    \hline
    \multicolumn{9}{l}{NS: No significant difference, *: $p<0.05$, **: $p<0.01$}
    \end{tabular}}
\end{table*}
\begin{table*}[tb]
    \centering
    \caption{The number of fitness evaluations (VRP). Boldface: minimum in column. Underline: minimum in row}
    \label{tab:eval_vrp}
    \scalebox{0.9}{
    \begin{tabular}{rl||rrrrrrrr!{\bvline{1pt}}rrrrrrrr}
    \hline
    &&\multicolumn{8}{c!{\bvline{1pt}}}{VRP1 ($\times 10^4$)}&\multicolumn{8}{c}{VRP2 ($\times 10^4$)}\\
      Mig.   &&\multicolumn{7}{c}{\# of utilized cores}&&\multicolumn{7}{c}{\# of utilized cores}&\\
        \cline{3-10}\cline{11-18}
     gap    &&  1&2&4&8&16&32&64&SD&  1&2&4&8&16&32&64&SD\\
    \hline
    16&median&{\textbf{56.1}}&{\textbf{55.6}}&{\textbf{58.2}}&{\textbf{55.7}}&{\textbf{54.1}}&{\textbf{53.5}}&{\textbf{49.5}}&NS&{\textbf{104}}&{\textbf{111}}&{\textbf{107}}&{{111}}&{\textbf{104}}&{\textbf{101}}&\underline{\textbf{99.9}}&*\\
    &stdv.&7.67&8.94&10.1&8.38&7.74&8.14&12.2&&9.15&14.3&27.7&24.9&17.7&12.4&18.5&\\
    \hline
    32&median&{{56.9}}&{{61.1}}&{{61.9}}&{{60.2}}&{{55.2}}&{{55.1}}&\underline{{50.6}}&**&{{105}}&{{113}}&{{117}}&{\textbf{110}}&{{113}}&\underline{{105}}&{{106}}&**\\
    &stdv.&6.53&7.61&9.85&8.26&8.52&8.11&9.11&&16.5&15.7&19.1&13.9&13.9&13.1&15.6&\\
    \hline
    64&median&{{60.1}}&{{61.0}}&{{59.1}}&{{59.5}}&{{54.6}}&{{57.0}}&\underline{{52.3}}&**&{{116}}&{{120}}&{{120}}&{{120}}&{{116}}&{{114}}&\underline{{108}}&*\\
    &stdv.&4.87&9.45&9.31&13.2&7.41&9.28&9.16&&15.1&13.0&16.5&13.5&12.4&16.8&17.9&\\
    \hline
    128&median&{{62.4}}&{{66.4}}&{{66.3}}&{{65.9}}&{{62.3}}&\underline{{57.9}}&{{58.2}}&**&{{119}}&{{126}}&{{125}}&{{124}}&{{119}}&{{113}}&\underline{{111}}&*\\
    &stdv.&7.68&9.72&10.2&12.1&9.41&10.0&10.5&&15.6&17.7&19.4&17.8&14.8&13.3&18.2&\\
    \hline
    256&median&{{69.7}}&{{68.8}}&{{68.3}}&{{66.6}}&{{68.4}}&{{63.6}}&\underline{{57.4}}&**&{{131}}&{{137}}&{{137}}&{{137}}&{{132}}&{{116}}&\underline{{115}}&**\\
    &stdv.&7.22&8.48&6.47&11.0&7.53&9.54&12.1&&12.1&17.7&18.3&13.4&15.6&16.2&14.0&\\
    \hline
    \hline
    &SD&**&**&**&**&**&**&*&&**&**&**&**&**&**&**&\\
    \hline
    \multicolumn{9}{l}{NS: No significant difference, *: $p<0.05$, **: $p<0.01$}\\
\end{tabular}}
\end{table*}
Tables~\ref{tab:eval_ppeaks} and \ref{tab:eval_vrp} show the number of fitness evaluations on P-PEAKS 20-100, P-PEAKS 200-100, VRP1, and VRP2. As in the previous tables, the results of the Kruskal-Wallis test are shown in the ``SD'' row and column. The smallest median values between the different migration gaps are indicated in boldface if there is a significant difference. The smallest ones between the different cores are indicated with an underlined text if a significant difference is found. 

For P-PEAKS 20-100, we found significant differences between the different migration gaps when using 32 or few cores, while no significant difference is found when using 64 cores. When the number of cores is below 32, the smaller migration gap obtains the smaller number of fitness evaluations. In particular, the migration gap of 16 achieves the best results when using 16 or fewer cores, while the migration gap of 32 is the best when using 32 cores. 

From P-PEAKS 200-1000, a significant difference between the difference migration gaps is found in all the number of cores on P-PEAKS 200-1000. When using one core, the migration gap of 16 obtains the smallest number of fitness evaluations. On the other hand, for more cores, the migration gap of 32 achieves the smallest number of fitness evaluations, and the fitness evaluations increase as the migration gap increases.

As to the different sized P-PEAKS problems, the number of fitness evaluations increases about ten times from P-PEAKS 20-100 to P-PEAKS 200-1000. Let us consider this carefully. Since the bit length of solutions is ten times longer in the largest problem, the search space is $2^{10}$ times larger, which is much more than ten times. Nevertheless, the impact on the numerical effort is only ten times larger. This indicates that PGA is very robust as a solver as problems grow. Such robustness (scalability) is also demonstrated in the literature of \cite{Alba2002} where a cellular model (fine-grained) PGA is used, and our result demonstrates the robustness of PGA even using the island model. As for the effect of the migration gap, it significantly impacts the numerical effort of both. It seems that a small migration gap leads to a reduced numerical effort.

The result on VRP1 shows a significant difference across the different migration gaps for all the utilized cores. The migration gap of 16 achieves the smallest number of fitness evaluations, and the smaller the migration gap, the smaller the number of fitness evaluations. The different number of cores throw different results with a significant difference, except for the migration gap of 16. When considering 32 or larger migration gaps, the smallest number of fitness evaluations is achieved using 64 cores. The migration gap of 16 also obtains the smallest number of fitness evaluations when using 64 cores though there is no significant difference.

We found that VRP1 and VRP2 follow similar trends: a significant difference between the different migration gaps is found for all cores. The migration gap of 16 shows the smallest number of fitness evaluations except for when using eight cores. Using eight cores, the migration gap of 32 indicates the best, and the migration gap of 16 shows the second best.
Overall, the smaller the migration gap, the smaller the number of fitness evaluations.

Smaller migration gaps can reduce the numerical effort for all problems and all number of used cores from these results. In particular, the migration gap of 16 achieved the best smallest number of fitness evaluations on most of our experiments. This happens because, in the problem instance used in our experiment, it was more effective to quickly share a better solution to search promising search region with a small migration gap than to maintain diversity between subpopulations and search independently with a large migration gap.

\begin{figure*}[!tb]
\begin{tabular}{cc}
\begin{minipage}[t]{0.49\linewidth}
    \centering
    \includegraphics[scale=0.6]{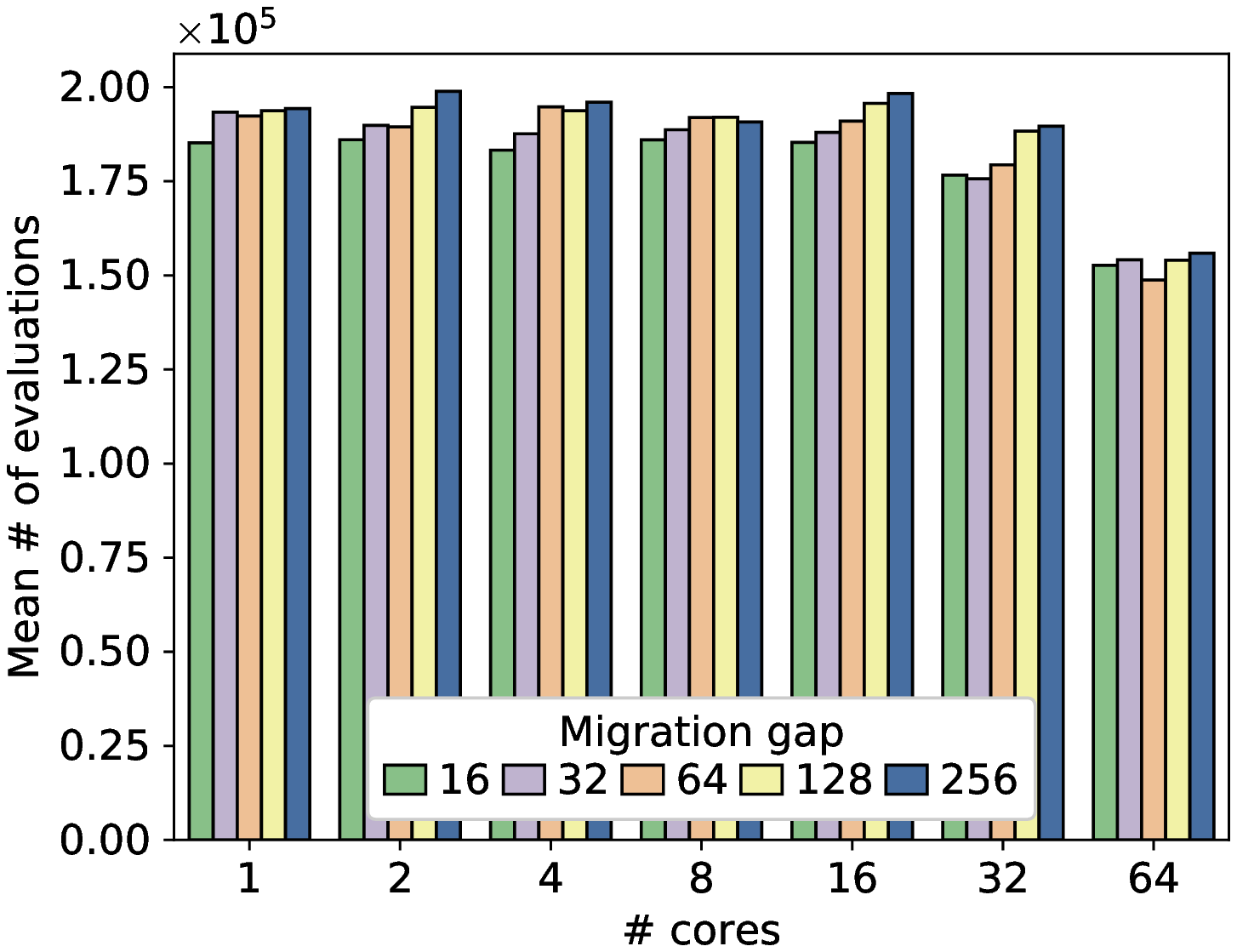}
    \caption{The median number of fitness evaluations on P-PEAKS 20-100}
    \label{fig:eval_med_exp1}
\end{minipage}&
\begin{minipage}[t]{0.49\linewidth}
    \centering
    \includegraphics[scale=0.6]{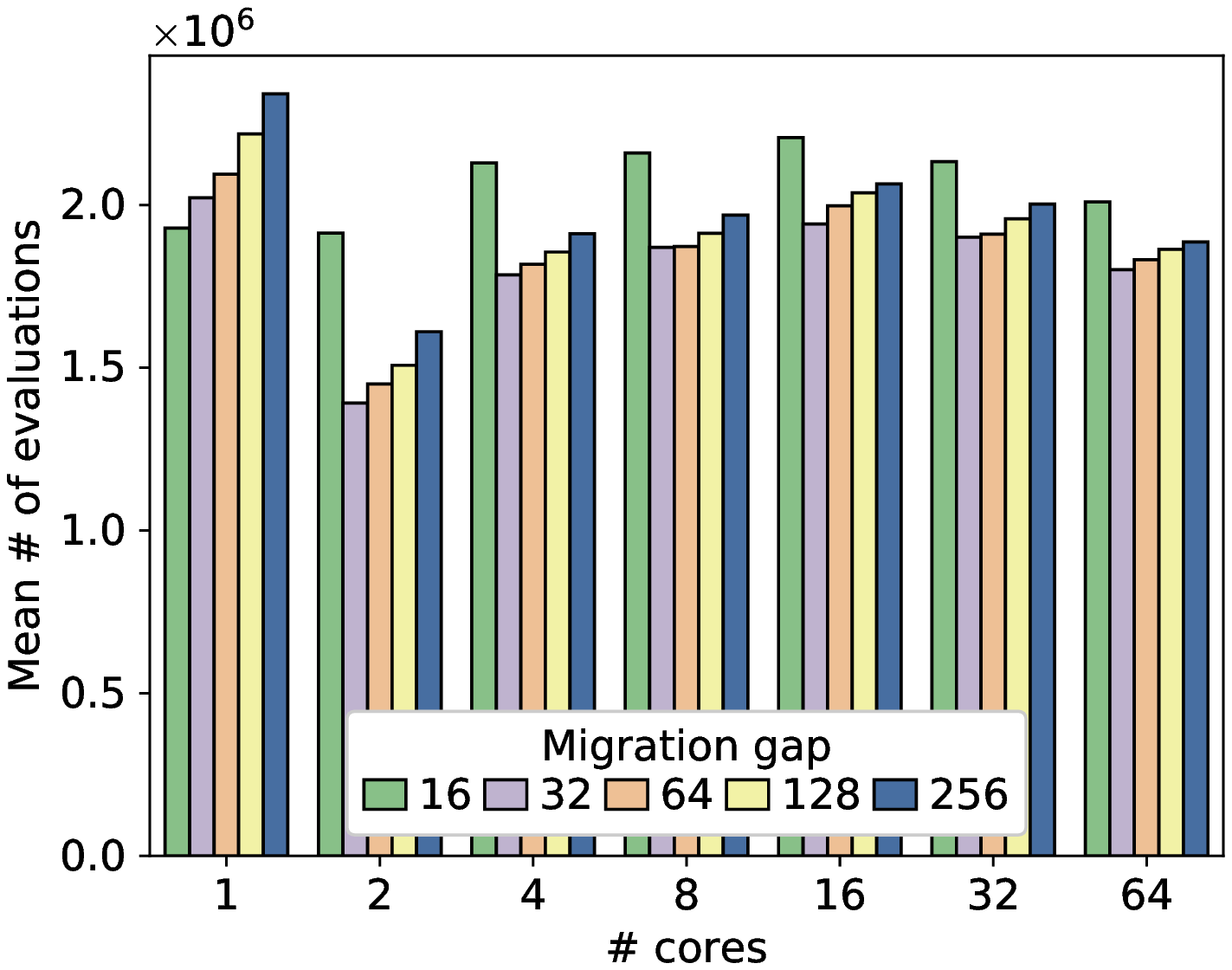}
    \caption{The median number of fitness evaluations on P-PEAKS 200-1000}
    \label{fig:eval_med_exp1_large}
\end{minipage}\\
\begin{minipage}[t]{0.49\linewidth}
    \centering
    \includegraphics[scale=0.6]{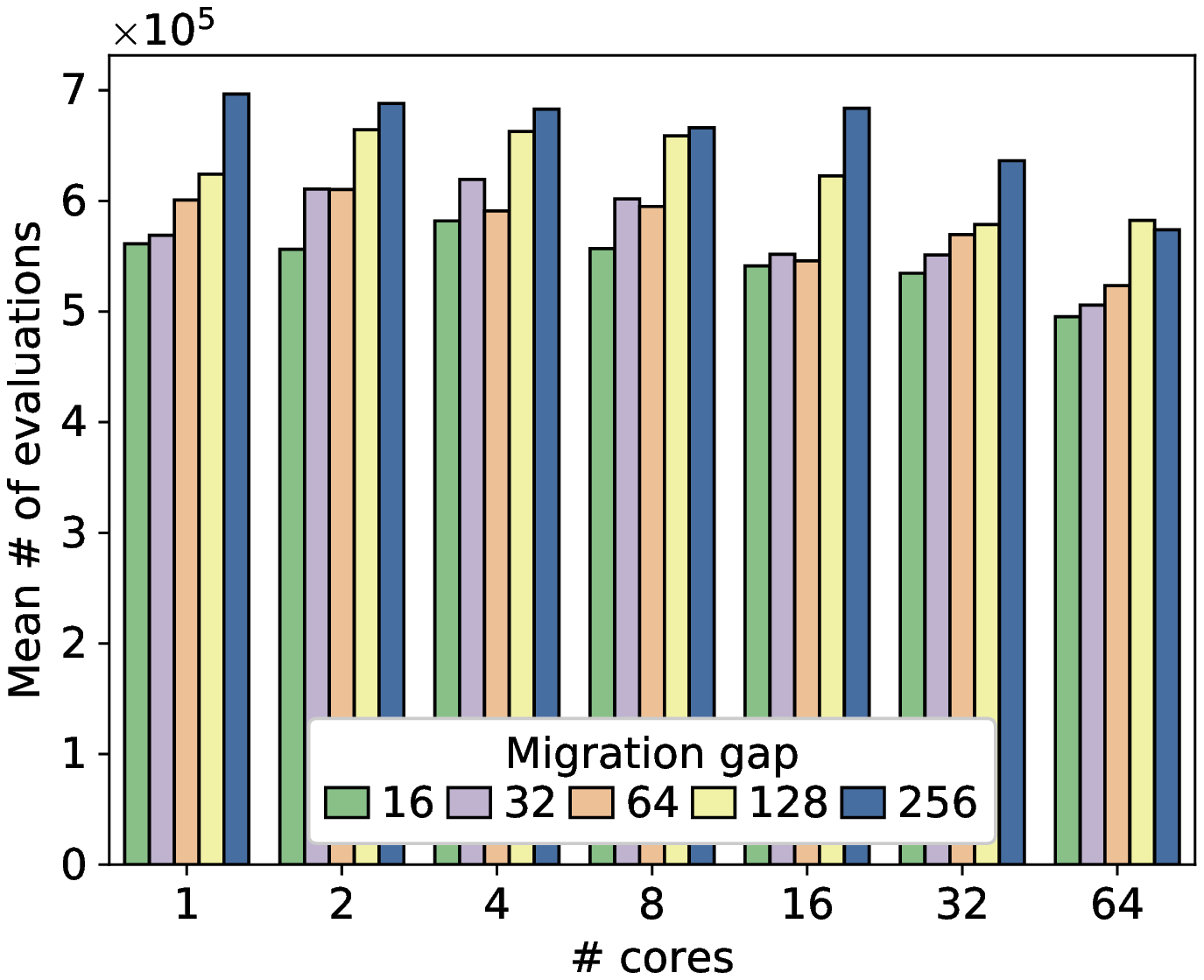}
    \caption{The median number of fitness evaluations on VRP1}
    \label{fig:eval_med_exp1_vrp1}
\end{minipage}&
\begin{minipage}[t]{0.49\linewidth}
    \centering
    \includegraphics[scale=0.6]{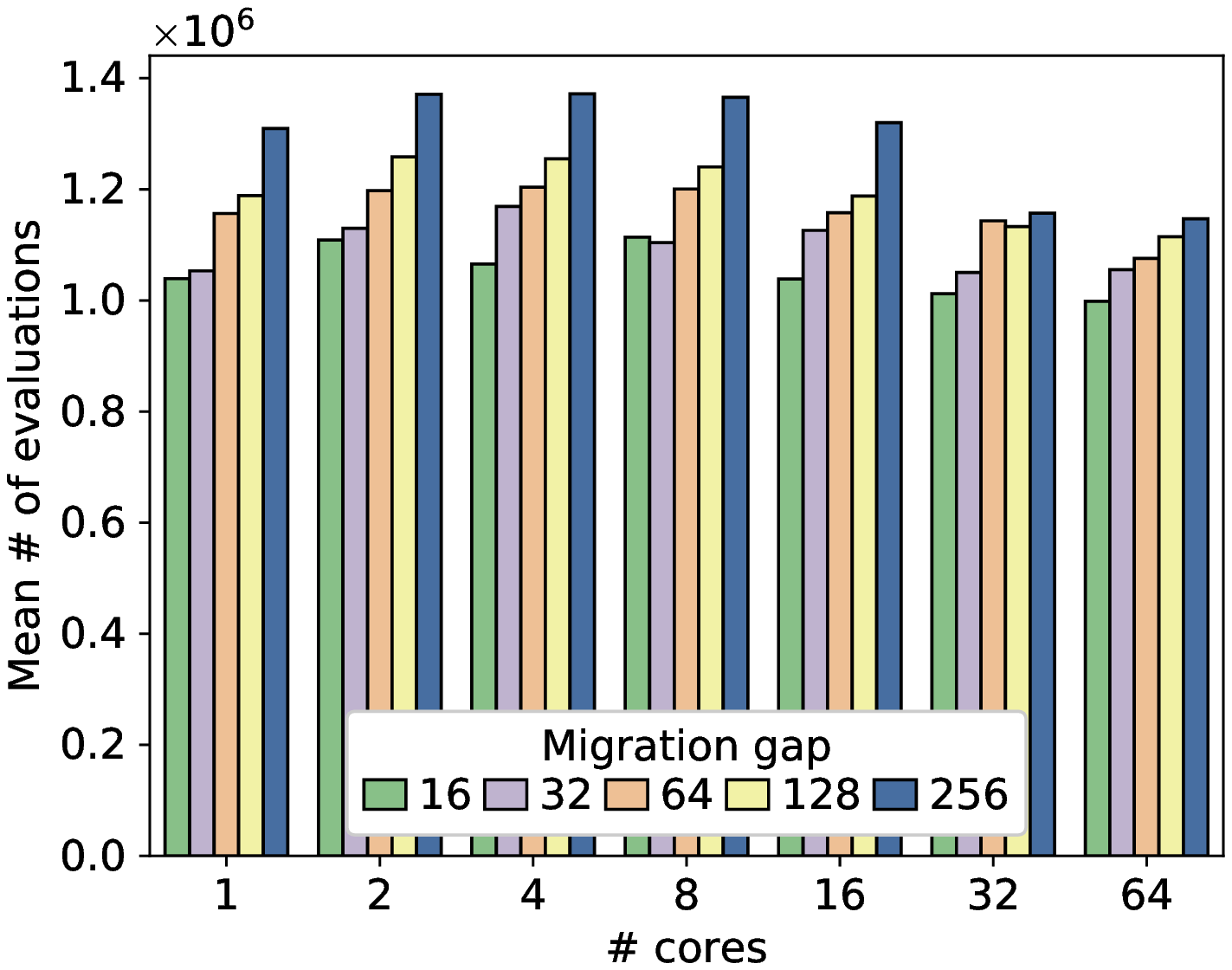}
    \caption{The median number of fitness evaluations on VRP2}
    \label{fig:eval_med_exp1_vrp2}
\end{minipage}
\end{tabular}
\end{figure*}
The number of cores used is an important factor for all problems and experiments, and it clearly states the importance of having more powerful machines and the potentially good results in later research with PGAs on these machines. On P-PEAKS 20-100, VRP1, and VRP2, the smallest number of fitness evaluations is achieved when using 64 cores (for any migration gap).
On the other hand, on P-PEAKS 200-1000, although a significant difference is found, the smallest number of fitness evaluations is achieved using two cores for all migration gaps.

\subsection{Understanding the Number of Fitness Evaluations}
This subsection shows some bar graphs of the number of fitness evaluations and discusses their tendency to ease the understanding of the previous tables.

Figs.~\ref{fig:eval_med_exp1}--\ref{fig:eval_med_exp1_vrp2} show the median number of fitness evaluations on P-PEAKS 20-100, P-PEAKS 200-1000, VRP1, and VRP2, respectively. 
The horizontal axis shows the number of used cores, while the vertical axis shows the median number of fitness evaluations. The different colors indicate the different migration gaps. 

From Fig.~\ref{fig:eval_med_exp1}, we can confirm that the number of fitness evaluations slightly increases as the migration gap increases when using 32 or fewer cores. When using 64 cores, such a tendency is not found. The number of fitness evaluations is smallest when the migration gap is 64, but no important difference is found.
By comparing the different number of utilized cores, 64 cores remarkably decreased the number of fitness evaluations with respect to the other cases.

In Fig.~\ref{fig:eval_med_exp1_large}, we can see that the number of fitness evaluations increases as the migration gap increases when using one core. On the other hand, when using two or more cores, the maximum number of fitness evaluations appeared when using a migration gap of 16. As to the rest of the migration gaps, 32 provokes the shortest number of fitness evaluations, and it increases as the migration gap increases. 

In Fig.~\ref{fig:eval_med_exp1_vrp1}, we notice that the number of fitness evaluations decreases with an also decreasing migration gap on VRP1. On the other hand, we can observe a trend to reduce the number of evaluations when the number of cores is increased. This difference, although it is statistically validated, is relatively small. This indicates that a different number of cores does not hugely impact the search behavior for this problem. However, it allows reducing the execution importantly, as we showed in the previous section.

A similar tendency to VRP1 can be found in VRP2 from Fig.~\ref{fig:eval_med_exp1_vrp2}.
When using the migration gap of 16, we got a minimum number of fitness evaluations. It increases with the migration gap. On the other hand, similar to the previous instance, the increment of the number of cores allows a small but statistically significant reduction of the number of evaluations, especially for the largest values of the migration gap.

As the overall conclusions of the numerical performance, we can clearly point to the importance of the migration gap for complex problems, in which a lower value can help faster converge to the best solution. However, in some cases (P-PEAKS 200-1000), a very low value of this parameter can mislead the search and delay reaching a high-quality solution. A problem-dependant trade-off seems then needed. The influence of the number of cores is less evident than that of the migration gap. The trend of the results shows that a higher number of cores will reduce the number of evaluations needed, but it strongly depends on the instance problem.

\section{Analysis of the Speed-up}\label{sec:su}
As we previously mentioned in Section~\ref{sec:eval_crit}, the speed-up is a fundamental metric for parallels techniques since it measures the gain in using additional physical computational resources. Analyzing its value for the different number of processor units is important to understand how the algorithm makes use of these additional hardware resources. It also allows a researcher to make some decisions about the used hardware platform since it informs the trade-off between the extra cost and the benefits of using more resources.

This section analyzes the speed-up performance by increasing the number of used cores and running on them always a single same algorithm. First, we calculate the speed-up from the median wall-clock time shown in Section~\ref{sec:time}. Then, we present the two fitting models and compare their parameters.

\subsection{Measuring Speed-up}
First, let us calculate the speed-up performance on each problem. In addition, we will perform a Kruskal-Wallis test to confirm the statistical difference between the different migration gaps and the number of used cores.

\begin{table*}[t]
    \centering
    \caption{Speed-up (P-PEAKS). Boldface: maximum in column. Underline: maximum in row. Italic: super-linear speed-up}
    \label{tab:speedup_ppeaks}
    \scalebox{0.75}{
    \begin{tabular}{rl||rrrrrrrr!{\bvline{1pt}}rrrrrrrr}
    \hline
    &&\multicolumn{8}{c!{\bvline{1pt}}}{P-PEAKS 20-100}&\multicolumn{8}{c}{P-PEAKS 200-1000}\\
      Mig.   &&\multicolumn{7}{c}{\# of utilized cores}&&\multicolumn{7}{c}{\# of utilized cores}&\\
        \cline{3-18}
     gap    &  &1&2&4&8&16&32&64&SD&1&2&4&8&16&32&64&SD\\
    \hline
    \rowcolor{gray}16&Real SU&{{{1.00}}}&{{\textit{2.02}}}&{{\textit{4.07}}}&{{{7.74}}}&{{{14.98}}}&{{{29.48}}}&\underline{{{49.12}}}&**&{{{1.00}}}&{{{1.95}}}&{{{3.71}}}&{{{7.10}}}&{{{13.66}}}&{{{26.71}}}&\underline{{{43.57}}}&**\\
    &Pred.&0.80&1.87&3.99&8.06&15.62&28.79&49.28&(MAE)&0.77&1.76&3.71&7.43&14.28&26.00&43.75&(MAE)\\
    &Pred. err.&+0.20&+0.15&+0.08&-0.31&-0.65&+0.69&-0.16&(0.32)&+0.23&+0.19&-0.00&-0.33&-0.62&+0.72&-0.18&(0.32)\\
    \hline
    \rowcolor{gray}32&Real SU&{{{1.00}}}&{\textbf{\textit{2.08}}}&{\textbf{\textit{4.16}}}&{{{7.94}}}&{\textbf{{15.48}}}&{\textbf{{30.97}}}&\underline{{{50.88}}}&**&{{{1.00}}}&{{{1.93}}}&{{{3.80}}}&{{{7.12}}}&{{{14.13}}}&{{{27.20}}}&\underline{{{44.14}}}&**\\
    &Pred.&0.72&1.86&4.09&8.38&16.32&30.03&51.11&(MAE)&0.73&1.76&3.77&7.60&14.62&26.53&44.31&(MAE)\\
    &Pred. err.&+0.28&+0.22&+0.07&-0.44&-0.84&+0.94&-0.23&(0.43)&+0.27&+0.17&+0.03&-0.48&-0.49&+0.68&-0.18&(0.33)\\
    \hline
    \rowcolor{gray}64&Real SU&{{{1.00}}}&{{\textit{2.06}}}&{{{3.98}}}&{{{7.75}}}&{{{15.18}}}&{{{30.17}}}&\underline{{{52.46}}}&**&{{{1.00}}}&{{{1.91}}}&{{{3.86}}}&{{{7.35}}}&{{{14.22}}}&{{{28.05}}}&\underline{{{45.59}}}&**\\
    &Pred.&0.82&1.88&3.97&8.03&15.72&29.61&52.58&(MAE)&0.71&1.76&3.81&7.73&14.94&27.23&45.79&(MAE)\\
    &Pred. err.&+0.18&+0.18&+0.02&-0.27&-0.54&+0.56&-0.12&(0.27)&+0.29&+0.15&+0.05&-0.38&-0.72&+0.82&-0.20&(0.37)\\
    \hline
    \rowcolor{gray}128&Real SU&{{{1.00}}}&{{\textit{2.03}}}&{{\textit{4.04}}}&{{{7.88}}}&{{{15.05}}}&{{{29.07}}}&\underline{{{51.41}}}&**&{{{1.00}}}&{{{1.96}}}&{{\textit{4.01}}}&{{{7.64}}}&{{{14.78}}}&{{{29.01}}}&\underline{{{47.51}}}&**\\
    &Pred.&0.97&1.99&3.99&7.90&15.33&28.84&51.45&(MAE)&0.75&1.83&3.94&7.99&15.45&28.25&47.69&(MAE)\\
    &Pred. err.&+0.03&+0.04&+0.05&-0.02&-0.28&+0.23&-0.04&(0.10)&+0.25&+0.14&+0.07&-0.35&-0.68&+0.76&-0.19&(0.35)\\
    \hline
    \rowcolor{gray}256&Real SU&{{{1.00}}}&{{{1.99}}}&{{{3.98}}}&{{{7.91}}}&{{{14.81}}}&{{{28.85}}}&\underline{{{50.70}}}&**&{{{1.00}}}&{\textbf{{1.98}}}&{\textbf{\textit{4.11}}}&{\textbf{{7.89}}}&{\textbf{{15.39}}}&{\textbf{{29.94}}}&\underline{\textbf{{49.74}}}&**\\
    &Pred.&0.95&1.96&3.96&7.84&15.21&28.56&50.76&(MAE)&0.76&1.87&4.04&8.22&15.94&29.30&49.90&(MAE)\\
    &Pred. err.&+0.05&+0.03&+0.03&+0.07&-0.40&+0.29&-0.06&(0.13)&+0.24&+0.10&+0.06&-0.33&-0.55&+0.63&-0.15&(0.30)\\
    \hline
    \hline
    &SD&NS&*&**&NS&*&**&NS&&NS&**&**&**&**&**&**&\\
    \hline
    \multicolumn{9}{l}{NS: No significant difference, *: $p<0.05$, **: $p<0.01$}
    \end{tabular}}
\end{table*}
\begin{table*}[!tb]
    \centering
    \caption{Speed-up (VRP). Boldface: maximum in column. Underline: maximum in row. Italic: super-linear speed-up}
    \label{tab:speedup_vrp}
    \scalebox{0.75}{
    \begin{tabular}{rl||rrrrrrrr!{\bvline{1pt}}rrrrrrrr}
    \hline
    &&\multicolumn{8}{c!{\bvline{1pt}}}{VRP1}&\multicolumn{8}{c}{VRP2}\\
      Mig.   &&\multicolumn{7}{c}{\# of utilized cores}&&\multicolumn{7}{c}{\# of utilized cores}&\\
        \cline{3-18}
     gap    &  &1&2&4&8&16&32&64&SD&1&2&4&8&16&32&64&SD\\
    \hline
    \rowcolor{gray}16&Real SU&{{{1.00}}}&{{{1.92}}}&{{{3.69}}}&{{{6.95}}}&{{{13.05}}}&{{{21.70}}}&\underline{{{26.77}}}&**&{{{1.00}}}&{{{1.85}}}&{{{3.71}}}&{{{6.60}}}&{{{13.06}}}&{{{22.43}}}&\underline{{{26.35}}}&**\\
    &Pred.&0.53&1.75&3.97&7.73&13.36&20.37&27.33&(MAE)&0.39&1.65&3.94&7.81&13.50&20.46&27.25&(MAE)\\
    &Pred. err.&+0.47&+0.17&-0.27&-0.79&-0.32&+1.34&-0.56&(0.56)&+0.61&+0.21&-0.23&-1.21&-0.45&+1.96&-0.89&(0.80)\\
    \hline
    \rowcolor{gray}32&Real SU&{{{1.00}}}&{{{1.80}}}&{{{3.43}}}&{{{6.51}}}&{{{12.83}}}&{{{20.90}}}&\underline{{{25.86}}}&**&{{{1.00}}}&{{{1.83}}}&{{{3.46}}}&{{{6.63}}}&{{{12.13}}}&{{{22.05}}}&\underline{{{25.37}}}&**\\
    &Pred.&0.45&1.63&3.78&7.44&12.90&19.69&26.44&(MAE)&0.39&1.60&3.81&7.53&13.03&19.76&26.35&(MAE)\\
    &Pred. err.&+0.55&+0.17&-0.36&-0.93&-0.07&+1.22&-0.58&(0.55)&+0.61&+0.24&-0.35&-0.90&-0.90&+2.29&-0.99&(0.90)\\
    \hline
    \rowcolor{gray}64&Real SU&{{{1.00}}}&{{{1.92}}}&{{{3.85}}}&{{{6.89}}}&{{{13.43}}}&{{{21.23}}}&\underline{{{26.10}}}&**&{{{1.00}}}&{{{1.91}}}&{{{3.71}}}&{{{6.79}}}&{{{12.92}}}&{{{22.39}}}&\underline{{{27.53}}}&**\\
    &Pred.&0.54&1.79&4.05&7.85&13.40&20.14&26.64&(MAE)&0.50&1.70&3.91&7.70&13.45&20.77&28.22&(MAE)\\
    &Pred. err.&+0.46&+0.13&-0.20&-0.96&+0.03&+1.09&-0.54&(0.49)&+0.50&+0.20&-0.20&-0.91&-0.53&+1.63&-0.69&(0.67)\\
    \hline
    \rowcolor{gray}128&Real SU&{{{1.00}}}&{{{1.84}}}&{{{3.58}}}&{{{6.53}}}&{{{12.45}}}&{{{21.89}}}&\underline{{{24.96}}}&**&{{{1.00}}}&{{{1.87}}}&{{{3.68}}}&{{{6.79}}}&{{{13.15}}}&{{{23.17}}}&\underline{{{28.02}}}&**\\
    &Pred.&0.38&1.62&3.88&7.63&13.11&19.67&25.96&(MAE)&0.40&1.64&3.93&7.85&13.77&21.25&28.84&(MAE)\\
    &Pred. err.&+0.62&+0.22&-0.30&-1.10&-0.65&+2.22&-1.01&(0.87)&+0.60&+0.23&-0.26&-1.05&-0.61&+1.92&-0.82&(0.79)\\
    \hline
    \rowcolor{gray}256&Real SU&{{{1.00}}}&{\textbf{{1.99}}}&{\textbf{{3.88}}}&{{{7.21}}}&{{{12.78}}}&{{{22.67}}}&\underline{{{28.10}}}&**&{{{1.00}}}&{{{1.92}}}&{{{3.75}}}&{{{6.90}}}&{{{13.21}}}&{\textbf{{25.33}}}&\underline{\textbf{{30.41}}}&**\\
    &Pred.&0.62&1.81&4.01&7.80&13.60&21.05&28.74&(MAE)&0.32&1.60&3.96&8.07&14.42&22.72&31.43&(MAE)\\
    &Pred. err.&+0.38&+0.18&-0.13&-0.59&-0.82&+1.62&-0.64&(0.62)&+0.68&+0.32&-0.21&-1.16&-1.21&+2.61&-1.03&(1.03)\\
    \hline
    \hline
    &SD&NS&*&**&NS&NS&NS&NS&&NS&NS&NS&NS&NS&**&**&\\
    \hline
    \multicolumn{9}{l}{NS: No significant difference, *: $p<0.05$, **: $p<0.01$}
    \end{tabular}}
\end{table*}
Tables~\ref{tab:speedup_ppeaks} and \ref{tab:speedup_vrp} show the speed-up calculated from the median wall-clock time when the algorithm is solving P-PEAKS 20-100, P-PEAKS 200-1000, VRP1, and VRP2, respectively.
The ``Real SU'' row indicates the real speed-up value calculated by Eq.~\eqref{eq:speed-up}. 
The largest speed-up values are indicated in boldface in these tables if there is a significant difference between the different migration gaps. The largest speed-up values in a row (across all number of cores) are underlined. Additionally, any results indicating super-linear speed-up (i.e., $S(m)>m$) are indicated in italics.

The remaining rows show the predicted results by the fitted models and the error between the real and the predicted speed-up values. The ``Pred.'' row indicates the predicted speed-up from the fitted rational function by Eq.~\eqref{eq:rat_model}, and the ``Pred. err.'' row shows the difference between the real speed-up and the predicted speed-up (real minus predicted).
At the ``SD'' column in the lines of the ``Pred. err.,'' the mean absolute error (MAE) is reported.

On P-PEAKS 20-100, we can find a significant difference between the different migration gaps when the number of utilized cores is 2, 4, 16, and 32. In contrast, no significant difference is found when using 8 and 64 cores. Especially when using 64 cores, there is no significant difference. This indicates that the migration gap does not significantly impact when using 64 cores for the small problem. As another interesting finding, when using 32 or fewer cores, the migration gap of 32 achieves the highest speed-up performance. It is worth noting that a super-linear speed-up is observed in some settings, i.e., the cases when using 2 and 4 cores.

On P-PEAKS 200-1000, a significant difference between the different migration gaps is found for any number of cores. In particular, the migration gap of 256 achieves the highest speed-up for all the number of used cores, and it decreases when using smaller migration gaps. When using four cores, the migration gaps of 128 and 256 show a super-linear speed-up.

A comparison of two P-PEAKS problems shows a different tendency. That is, a smaller migration gap obtains the better speed-up performance on P-PEAKS 20-100. In particular, the migration gap of 32 achieves the best speed-up when a significant difference is found. On the other hand, on P-PEAKS 200-1000, larger migration gaps, particularly the migration gap of 256, obtain better speed-up performance. This indicates that the migration gap should be selected depending on the problem size to achieve the best speed-up performance, and that it effectively affects speed-up.

From the result on VRP1, a significant difference between the different migration gaps is found when using 2 and 4 cores. In contrast, no significant difference is found for the other number of used cores. When using 2 and 4 cores, the migration gap of 256 shows the highest speed-up.

On VRP2, we can find a significant difference between the different migration gaps when using 32 and 64 cores, while no significant difference is found when using 16 or fewer cores. On VRP2, the larger the migration gap is used, the higher speed-up is achieved. That could be useful to know when approaching VRP problems with PGAs in the future.

The super-linear speed-up is found on P-PEAKS 20-100 and P-PEAKS 200-1000 when using 2 and 4 cores in the experiment. Both physical and algorithmic reasons can cause this. From the physical aspect, when the algorithm run is equally distributed among several cores, the memory can be more efficiently managed, e.g., the cache memory can be utilized because the memory needed in each core decreases. On the other hand, many cores imply an additional overhead mainly due to the communications. Therefore, using 2 or 4 cores may achieve a good trade-off between the communication overhead and memory utilization, and some of these cases produce the super-linear speed-up.

On the other hand, from the algorithmic aspect, asynchronous communication used in the experiment may affect the behavior of the speed-up. The asynchronous communication allows each process to run the algorithm independently, which is more beneficial to the search than running it on a single core. In P-PEAKS 200-1000, asynchronous communication has a positive effect, and the number of evaluations is significantly reduced, as shown in the previous section, leading to the super-linear speed-up.

A significant difference is found for all migration gaps across all number of cores, and the highest speed-up is achieved when using 64 cores for all problems and all migration gaps.

\begin{figure*}[bt]
\begin{tabular}{cc}
\begin{minipage}[t]{0.49\linewidth}
    \centering
    \includegraphics[scale=0.6]{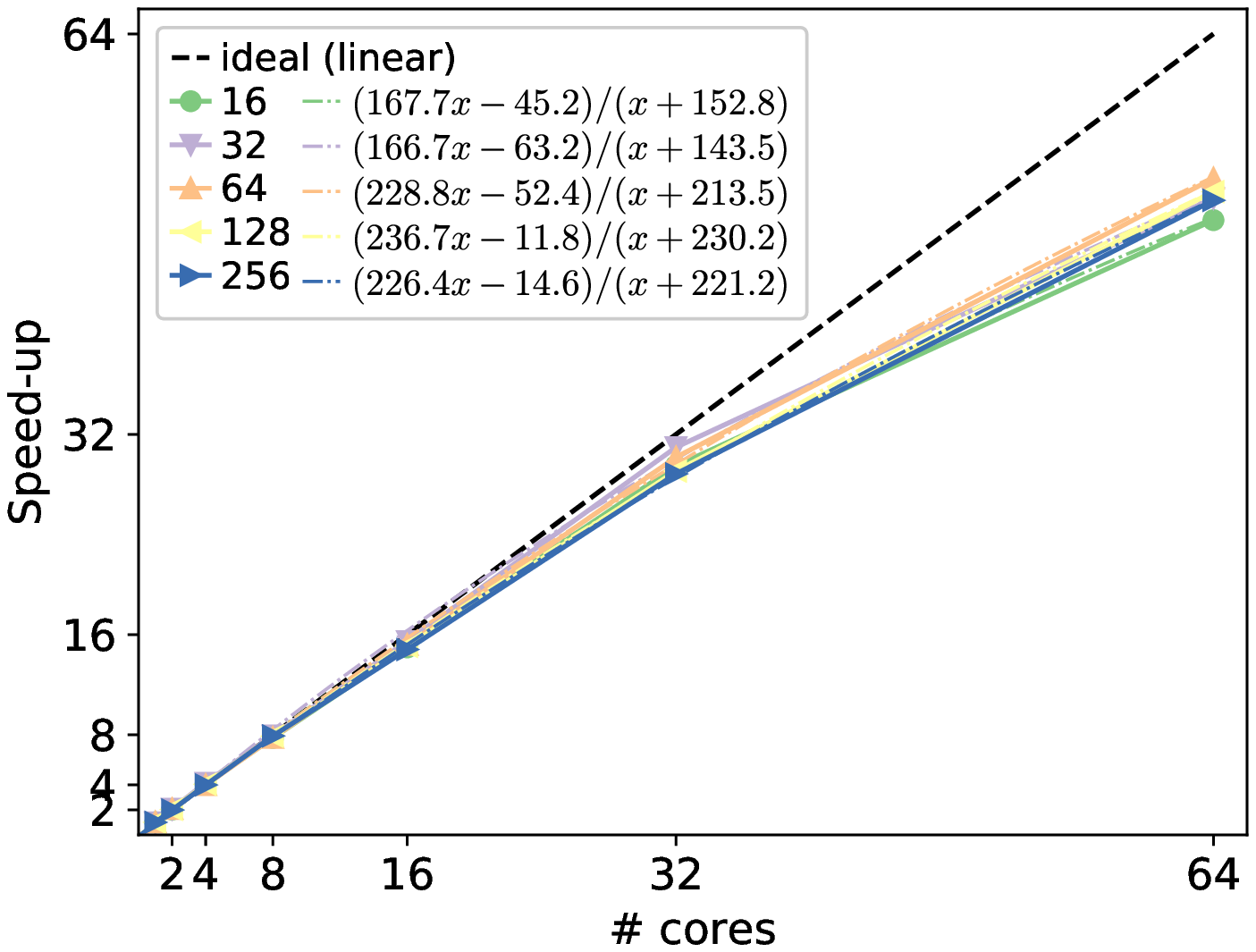}
    \caption{Speed-ups for each migration gap (P-PEAKS 20-100)}
    \label{fig:speedups_med_exp1}
\end{minipage}&
\begin{minipage}[t]{0.49\linewidth}
    \centering
    \includegraphics[scale=0.6]{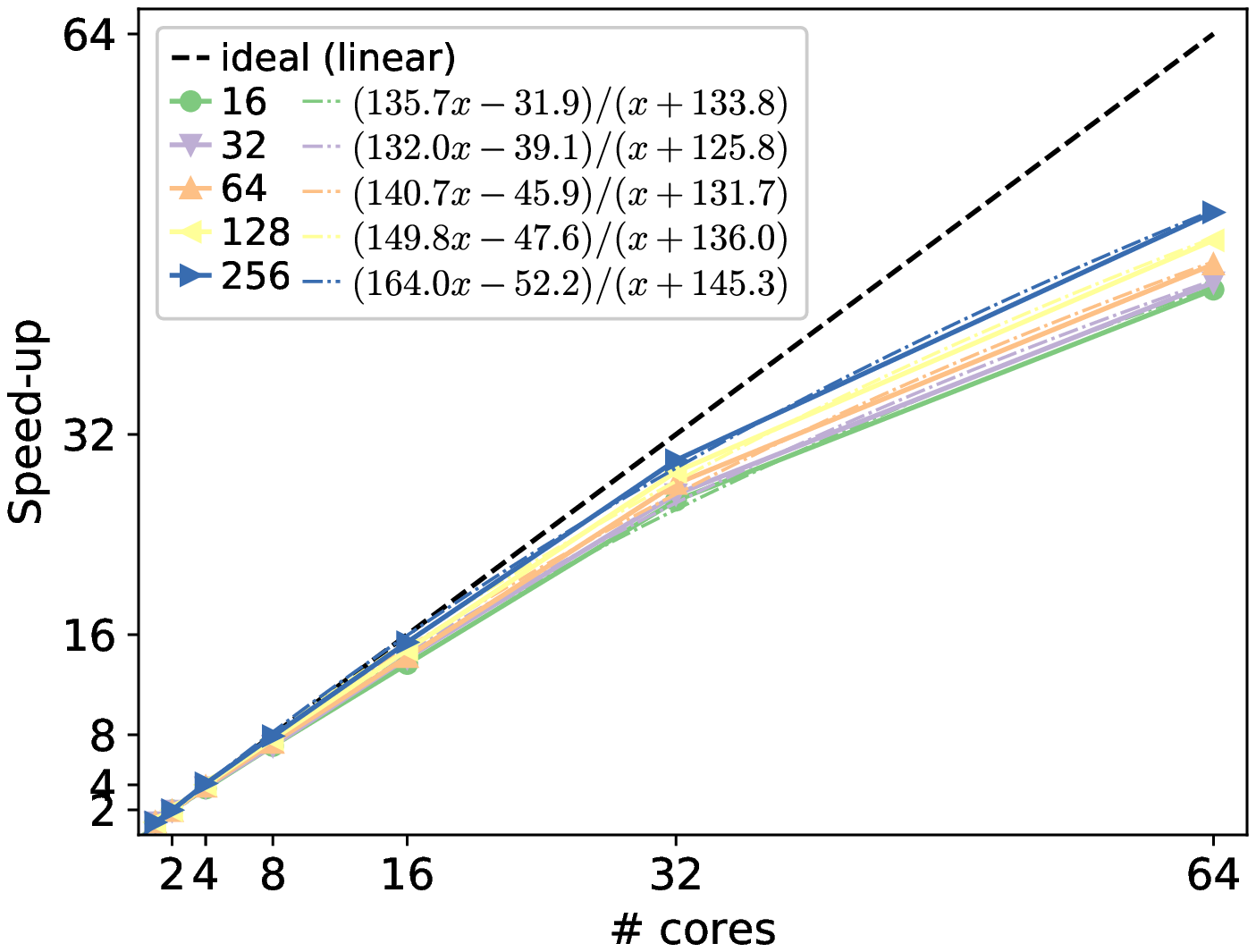}
    \caption{Speed-ups for each migration gap (P-PEAKS 200-1000)}
    \label{fig:speedups_med_exp1_large}
\end{minipage}\\
\begin{minipage}[t]{0.49\linewidth}
    \centering
    \includegraphics[scale=0.6]{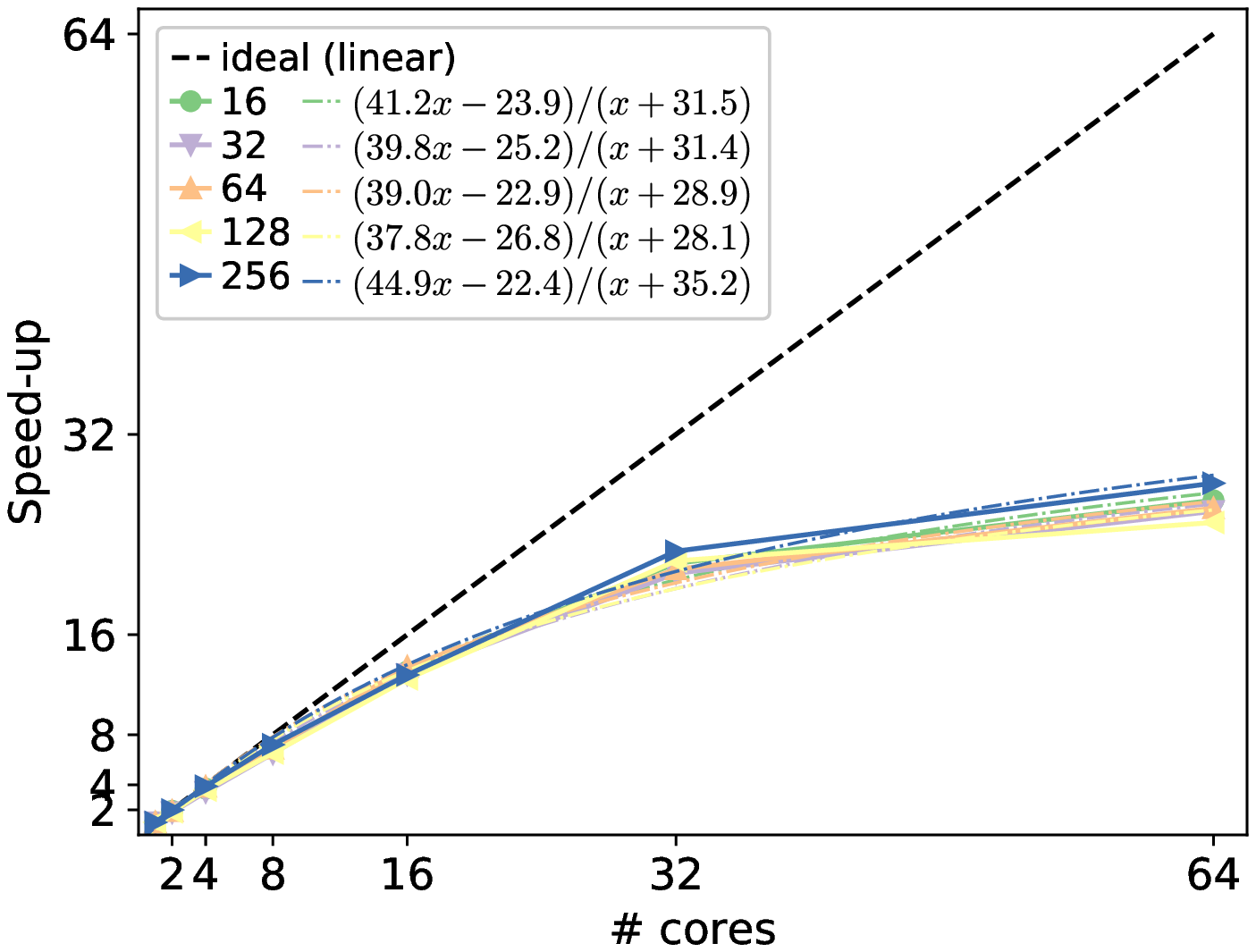}
    \caption{Speed-ups for each migration gap (VRP1)}
    \label{fig:speedups_exp1_vrp1}
\end{minipage}&
\begin{minipage}[t]{0.49\linewidth}
    \centering
    \includegraphics[scale=0.6]{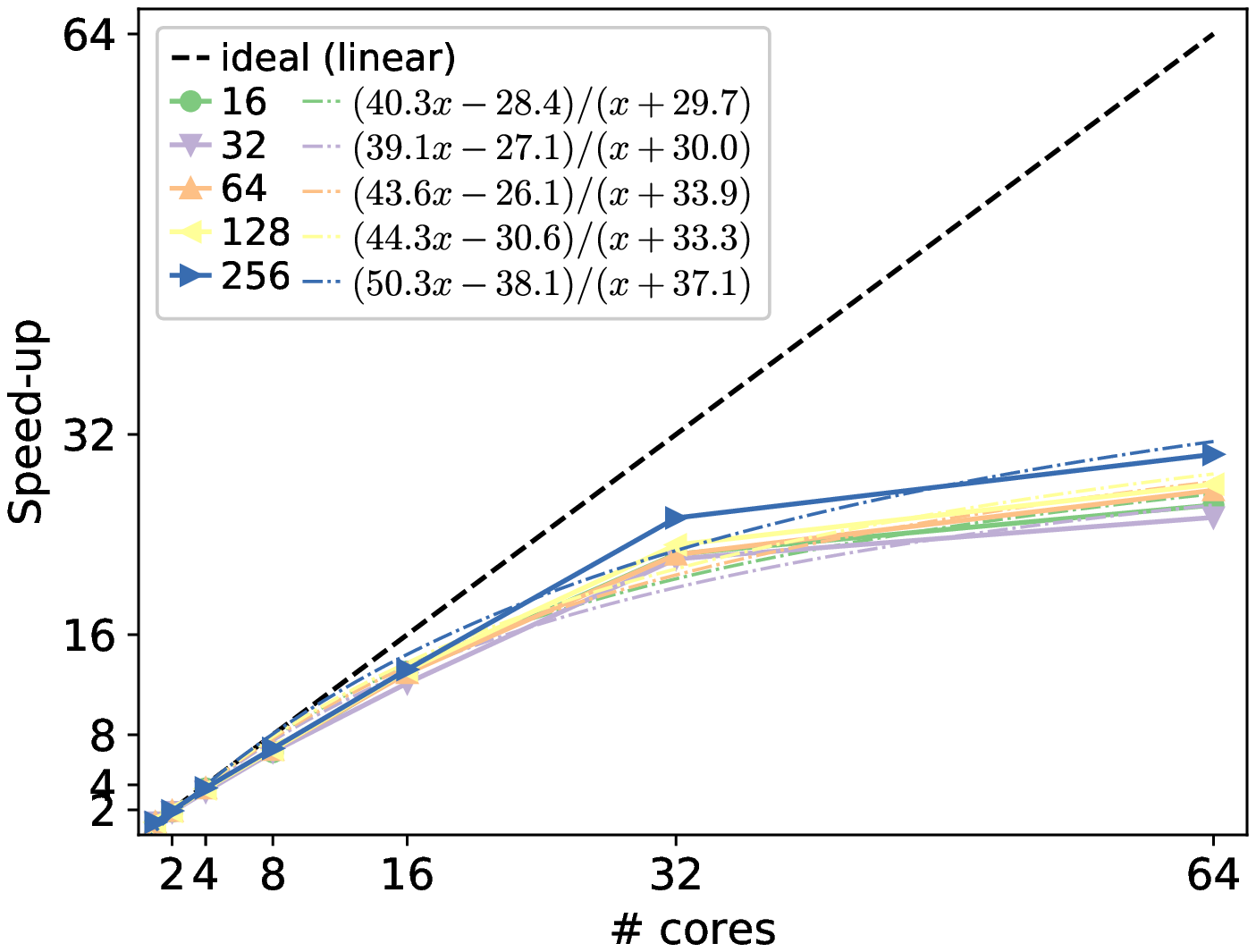}
    \caption{Speed-ups for each migration gap (VRP2)}
    \label{fig:speedups_exp1_vrp2}
\end{minipage}
\end{tabular}
\end{figure*}
As to the predicted model, the mean absolute error of the rational function model is very small. For the rational function model, the mean absolute error is less than 1.0 in the P-PEAKS problems and VRPs except for VRP2 with the migration gap of 256. In VRP2, the mean absolute error is 1.03 when using the migration gap of 256, but this prediction error is still small.

\subsection{Understanding Speed-up}\label{sec:understand_su}
In this subsection, we display the relation between the speed-up values and the number of used cores. Then, the fitted model with the rational function is computed and discussed. 

Figs.~\ref{fig:speedups_med_exp1}--\ref{fig:speedups_exp1_vrp2} show the speed-up for each migration gap and each number of used cores on P-PEAKS 20-100, P-PEAKS 200-1000, VRP1, and VRP2. The horizontal axis shows the number of used cores, while the vertical axis shows the speed-up calculated with the median wall-clock time. The solid lines connect the real speed-ups. The dash-dotted lines show the fitted lines with the rational function model. The black dashed line shows the ideal (linear) speed-up where $S(m)=m$. The predicted equations are denoted in the legend.

Fig.~\ref{fig:speedups_med_exp1} shows the result on P-PEAKS 20-100.
The fitted lines for each migration gap are similar, so the migration gap does not highly affect speed-up here. 

Fig.~\ref{fig:speedups_med_exp1_large} shows the result on P-PEAKS 200-1000.
From this figure, as similar to P-PEAKS 20-100, the rational model seems to be well-fitted to the real speed-up. Comparing the different migration gaps, we can notice that the predicted speed-up increases with the migration gap, being 256 the one producing the largest predicted speed-up.

From Fig.~\ref{fig:speedups_exp1_vrp1}, on VRP1, the fitted lines of the rational function are well-fitted to the real speed-up. The fitted lines for each migration gap are similar, but the migration gap of 256 still shows slightly better speed-up performance than the others. 

From Fig.~\ref{fig:speedups_exp1_vrp2}, on VRP2, the fitted lines of the rational function are also well-fitted to the real speed-up. Again, the migration gap of 256 shows the best speed-up behavior compared with the other migration gaps from the fitted lines.

Similar to the analysis for the wall-clock time, we can conclude that parameter $a$ is the most relevant one when the number of cores is high. However, now for the speed-up, the importance of $b$ and $c$ is the opposite of the wall-clock time. Here, the influence of the parameter $b$ is negligible when the number of cores is larger than $2$, and the parameter $c$ is quite important when the number of cores is low.

\begin{table}[!tb]
\centering
\caption{Limit values of the predicted speed-up}
\label{tab:limit_values}
\scalebox{0.8}{
\small
\begin{tabular}{r|l|r|rrrr}
\hline
Mig.&&&\multicolumn{4}{c}{Ratio ($row / column$)}\\
gap&Problem&Limit&P-100&P-1000&VRP1&VRP2\\
\hline
16&P-PEAKS 20-100&167.68&1.00&1.24&4.07&4.16\\
&P-PEAKS 200-1000&135.70&0.81&1.00&3.30&3.36\\
&VRP1&41.15&0.25&0.30&1.00&1.02\\
&VRP2&40.33&0.24&0.30&0.98&1.00\\
\hline
32&P-PEAKS 20-100&166.66&1.00&1.26&4.19&4.26\\
&P-PEAKS 200-1000&132.04&0.79&1.00&3.32&3.38\\
&VRP1&39.81&0.24&0.30&1.00&1.02\\
&VRP2&39.12&0.23&0.30&0.98&1.00\\
\hline
64&P-PEAKS 20-100&228.77&1.00&1.63&5.86&5.25\\
&P-PEAKS 200-1000&140.71&0.62&1.00&3.61&3.23\\
&VRP1&39.01&0.17&0.28&1.00&0.90\\
&VRP2&43.55&0.19&0.31&1.12&1.00\\
\hline
128&P-PEAKS 20-100&236.69&1.00&1.58&6.27&5.34\\
&P-PEAKS 200-1000&149.78&0.63&1.00&3.97&3.38\\
&VRP1&37.77&0.16&0.25&1.00&0.85\\
&VRP2&44.31&0.19&0.30&1.17&1.00\\
\hline
256&P-PEAKS 20-100&226.40&1.00&1.38&5.04&4.50\\
&P-PEAKS 200-1000&163.97&0.72&1.00&3.65&3.26\\
&VRP1&44.88&0.20&0.27&1.00&0.89\\
&VRP2&50.27&0.22&0.31&1.12&1.00\\
\hline
\end{tabular}
}
\end{table}
Table~\ref{tab:limit_values} summarizes the limit values of the predicted speed-up model of the rational function, i.e., $a$ in Eq.~\eqref{eq:rat_model}. The ``Limit'' column indicates the limit values of the fitted rational function where the number of used cores approaches infinity. The ``Ratio'' group indicates the ratio of limit values of different problems. Each value is calculated by dividing the gradient in the row by that in the column.

From Table~\ref{tab:limit_values}, on the P-PEAKS 20-100 problem, high speed-up performance is predicted. When using the migration gap of 128, the maximum 236.69 times faster execution is predicted than the single-core execution by the proposed model. 
On the P-PEAKS 200-1000 problem, the predicted speed-up is 62--81\% lower than P-PEAKS 20-100. However, comparing different migration gaps, the migration gap of 256 obtains the highest predicted speed-up of the maximum 163.97.

On the other hand, the predicted speed-up performance is limited on VRP1 and VRP2. On VRP1, the predicted speed-up limit values are up to 44.88, which is achieved with the migration gap of 256.
On VRP2, the speed-up limit value on VRPs is at most 50.27 obtained with the migration gap of 256.
Our model indicates that the predicted limit speed-up of VRPs is more than three times lower than that of the P-PEAKS problems. Whether this is a numerical indication of the difficulty in solving problems with PGA remains an interesting open issue.

The analysis using our mathematical model shows more potentialities than the results in Table~\ref{tab:speedup_ppeaks}, which simply compare figures of the speed-ups. In particular, the speed-up analysis does not show a significant difference when using 64 cores in P-PEAKS 20-100. However, the analysis using our model can show that there can be a large difference in speed-ups between different migration gaps when using more cores. Specifically, we can see that although the migration gap of 64 shows the highest speed-up when using 64 cores, our model concludes that the migration gap of 128 is the most promising when more cores are used. This indicates that the proposed model can provide insights into performance differences that cannot be found by only analyzing a limited number of cores (64 cores in our experiments), such as P-PEAKS 20-100.

\section{Summary\label{sec:summ}}
\begin{table*}[!tbp]
\newcolumntype{R}{>{\raggedright\arraybackslash}X}
\centering
\caption{Summary of the experiments}
\label{tab:summary}
\small
\begin{tabularx}{\textwidth}{l|p{0.13\columnwidth}|R|R|R}
\hline
     Viewpoint&Problem&Computational effort&Numerical effort&Speed-up  \\
\hline
\hline
    \multirow{4}{*}{\begin{minipage}{0.13\columnwidth}Migration gaps\end{minipage}}&P-PEAKS 20-100&The migration gap does not have a significant impact on the wall-clock time.&A smaller migration gap obtains a smaller number of fitness evaluations, but no difference between the migration gaps is found when using 64 cores.&The migration gap of 64 produces a higher speed-up, but there is no difference between the migration gaps when using 64 cores.  \\
    \cline{2-5}
    &P-PEAKS 200-1000&A smaller migration gap obtains shorter wall-clock time, but no difference between migration gaps when using 64 cores.&The migration gap of 32 brings the smallest number of fitness evaluations except for one core. When using one core, the migration gap of 16 brings the smallest number of fitness evaluations.&A larger migration gap obtains a higher speed-up. In particular, the migration gap of 256 produces the highest speed-up in all the number of used cores.\\
    \cline{2-5}
    &VRP1&A smaller migration gap obtains a shorter wall-clock time, but there is no difference between migration gaps when using 32+ cores.&A smaller migration gap produces a smaller number of fitness evaluations. In particular, the migration gap of 16 obtains the smallest number of fitness evaluations.&The migration gap does not have a significant impact on the speed-up performance when using 8+ cores. However, when using 2 and 4 cores, the migration gap of 256 obtains the highest speed-up. \\
    \cline{2-5}
    &VRP2&A smaller migration gap obtains a shorter wall-clock time, but there is no difference between migration gaps when using 64 cores.&A smaller migration gap produces a smaller number of fitness evaluations. In particular, the migration gap of 16 obtains the smallest number of fitness evaluations.&A larger migration gap achieves a higher speed-up when using 32+ cores. In particular, the migration gap of 256 obtains the highest speed-up. \\
\hline
\hline
    \multirow{4}{*}{\begin{minipage}{0.13\columnwidth}The number of cores\end{minipage}}&P-PEAKS 20-100 &The use of 64 cores produces the shortest wall-clock time for all migration gaps. &The use of 64 cores produces the smallest number of fitness evaluations for all migration gaps. &The use of 64 cores produces the highest speed-up for all migration gaps.  \\
    \cline{2-5}
    &P-PEAKS 200-1000   &The use of 64 cores produces the shortest wall-clock time for all migration gaps. &The use of two cores produces the smallest number of fitness evaluations for all migration gaps. &The use of 64 cores produces the highest speed-up for all migration gaps.\\
    \cline{2-5}
    &VRP1   &The use of 64 cores produces the shortest wall-clock time for all migration gaps. &The use of 64 cores produces the smallest number of fitness evaluations except for the migration gaps of 16 and 128. When using the migration gap of 16, the number of used cores does not significantly impact the number of fitness evaluations. When using the migration gap of 128, the use of 32 cores achieves the smallest number of fitness evaluations. &The use of 64 cores produces the highest speed-up for all migration gaps.  \\
    \cline{2-5}
    &VRP2   &The use of 64 cores produces the shortest wall-clock time for all migration gaps. &The use of 64 cores produces the smallest number of fitness evaluations for all migration gaps except for the migration gap of 32. When using the migration gap of 32, the use of 32 cores produces the smallest number of fitness evaluations.    &The use of 64 cores produces the highest speed-up for all migration gaps.  \\
\hline
\end{tabularx}
\end{table*}
This section summarizes the overall implications given by the experiments.
Table~\ref{tab:summary} shows a brief description of the experiments from the viewpoints of the migration gap and the number of used cores.

As to the migration gap, a small value seems to be the most effective way to reduce the computational effort, but the impact is negligible when using a large number of cores. 
For the numerical effort, a smaller migration gap is also effective in all benchmark problems. In particular, the migration gap of 16 obtains the smallest number of fitness evaluations in most cases of P-PEAKS 20-100, VRP1, and VRP2. From the speed-up viewpoint, the migration gap does not significantly impact P-PEAKS 20-100. Meanwhile, the larger migration gap, e.g., the migration gap of 256, produces the highest speed-up on the other problems.

The larger the number of cores used for running the island-based PGA the higher its performance, both in small computational effort and large speed-up. This seems common sense, but please remember that the number of islands is constant throughout all this study (i.e., we do not add more islands when more cores are used). It means this algorithm is really benefiting from parallelism and can scale for other problems and larger hardware settings.
As to the numerical effort, the more cores, the smaller the number of fitness evaluations. However, on P-PEAKS 200-1000, the smallest number of fitness evaluations is achieved when using only two cores, so exceptions clearly exist.

\section{Conclusion and Future Works\label{sec:conclu}}
In this article, we have tried to address an analysis of the parameters of the PGA (migration gap), the benefits of using a cluster of multi-core computers, and how all this relates to the problem solved (binary and permutation-based ones) at the same time (kind of holistic analysis of island PGA). We did it both in a descriptive manner, getting conclusions on all that, but also in a compact mathematical way, by summarizing our outcomes in simple numerical models to foster future research.

From the experimental results, some important conclusions can be drawn regarding \textbf{RQ1}. First, adding more hardware resources clearly allows, not only to reduce the wall-clock time, but also to impact on the numerical performance, i.e., a reduction of the number of evaluations. Second, the migration parameters also have a high importance in these two metrics. In this study, we have considered the migration gap, which determines how much coupled the islands are. 

For the tested instance problems (\textbf{RQ3}), a lower value (highly coupled islands) provides the most beneficial search properties, especially when the hardware resources are limited (less than 64 cores). 
An interesting finding of the computational effort is that the difference in the average evaluation time of the solution is directly reflected in the difference in execution times for VRPs of different sizes. This suggests that it may be possible to predict the execution time by regarding problem size differences, for the same class of problem instances.

With respect to the speed-up, our results show that it highly depends on the features of the problem, and while P-PEAKS obtains large gains as more resources are added (nearly linear, even with the high number of cores), VRP only obtains small speed-up improvements when more than 32 cores are used. It is clear that each type of problem has a range of computational resources worth using, which suggests interesting types of research in this sense in the future.

With respect to \textbf{RQ2}, we have also provided a simple mathematical model for the wall-clock times and the speed-up. These models allow studying these metrics in an easy way and beyond the intuitive look-at-the-graph traditional way, which is hard to compare and exploit in future studies. Using our model, only three parameters should be analyzed, and in many cases (e.g., with a medium/high number of cores), we can focus on only a single parameter ($a$). Using this later value, we can compare different factors and problems as we did in Table~\ref{tab:limit_values} and get sound conclusions that are easy to grasp.

We already mentioned some future work, and more is to come, e.g., in extending this analysis to the rest of the migration policy parameters, mainly the migration selection and the topology. In addition, increasing our benchmark problems with others (having different features) such as continuous problems or black-box functions (e.g., simulators), is expected to provide a more comprehensive study.

\appendix
\section{Model Selection}\label{sec:app_fun}
This appendix shows how we made the choice of the mathematical model used to understand the wall-clock time and the speed-up. First, we show the numerical functions we considered in making this paper (there are uncountable families of curves, we selected simple but accurate ones). Then, we show the result of model fitting for the wall-clock time and the speed-up and choose the best trade-off model for our analysis in terms of simplicity and accuracy.

\subsection{Used Functions}
In this subsection, we discuss the mathematical functions considered in this article. There are a lot of function families for fitting any actual curve. We have tested three function families for this study: linear, rational, and exponential/logarithm ones attending to the results and previous works. In particular, for modeling the wall-clock time and the speed-up, we consider the following six mathematical models:
\begin{itemize}
\item Linear function (Linear)
\begin{equation}
    f(x)=ax+b
    \label{eq:linear}
\end{equation}
\item Rational function with two parameters (Rational 1)
\begin{equation}
    f(x)=a+\frac{b}{x}
    \label{eq:rational1_wct}
\end{equation}
\item Rational function with two parameters (Rational 2)
\begin{equation}
    f(x)=\frac{ax}{x+b}
    \label{eq:rational2_wct}
\end{equation}
\item Rational function with three parameters (Rational 3)
\begin{equation}
    f(x)=\frac{ax+b}{x+c}
    \label{eq:rational3_wct}
\end{equation}
\item Exponential function with three parameters (Exponential)
\begin{equation}
    f(x)=a+be^{-cx}
    \label{eq:exponential_wct}
\end{equation}
\item Logarithmic function with two parameters (Logarithmic)
\begin{equation}
    f(x)=a+b\ln(x)
    \label{eq:logarithmic_wct}
\end{equation}
\end{itemize}
For each function, $x$ indicates the number of used cores, while $f(x)$ indicates the predicted response value, i.e., the wall-clock time or the speed-up.

\subsection{Model Selection for the Wall-clock Time}
In this subsection, we show the result of model fitting for the wall-clock time.
As the data for model fitting, we use the median wall-clock time between migration gaps shown in Section~\ref{sec:time}.

\begin{table*}[!htb]
    \centering
    \caption{Comparison of the predicted wall-clock time and its error from the real wall-clock time (P-PEAKS)}
    \label{tab:wct_model_comp_ppeaks}
    \scalebox{0.75}{
    \begin{tabular}{ll||rrrrrrrr!{\bvline{1pt}}rrrrrrrr}
    \hline
    &&\multicolumn{8}{c!{\bvline{1pt}}}{P-PEAKS 20-100}&\multicolumn{8}{c}{P-PEAKS 200-1000 $(\times10^3)$}\\
      &&\multicolumn{7}{c}{\# of utilized cores}&&\multicolumn{7}{c}{\# of utilized cores}&\\
        \cline{3-18}
     &&  1&2&4&8&16&32&64&MAE&  1&2&4&8&16&32&64&MAE\\
    \hline
    \rowcolor{gray}Real WCT&&6.21&3.05&1.53&0.79&0.41&0.21&0.12&&6.47&3.39&1.69&0.89&0.46&0.23&0.14&\\
    \hline
    Linear&Pred.&2.73&2.67&2.56&2.33&1.88&0.98&-0.82&&2.93&2.87&2.75&2.51&2.03&1.06&-0.86&\\
    &Error&+3.48&+0.38&-1.02&-1.54&-1.47&-0.77&+0.94&1.37&+3.55&+0.52&-1.06&-1.62&-1.57&-0.83&+1.00&1.45\\
    \hline
    Rational 1&Pred.&6.18&3.10&1.55&0.78&0.40&0.20&0.11&&6.52&3.29&1.68&0.87&0.47&0.27&0.17&\\
    &Error&+0.03&-0.04&-0.02&+0.01&+0.01&+0.00&+0.01&0.02&-0.05&+0.09&+0.01&+0.02&-0.01&-0.03&-0.02&0.03\\
    \hline
    Rational 2&Pred.&6.26&1.64&1.20&1.06&1.00&0.97&0.96&&6.53&1.81&1.33&1.17&1.11&1.08&1.06&\\
    &Error&-0.05&+1.41&+0.33&-0.26&-0.59&-0.76&-0.84&0.61&-0.06&+1.58&+0.36&-0.28&-0.65&-0.84&-0.92&0.67\\
    \hline
    Rational 3&Pred.&6.21&3.06&1.53&0.78&0.41&0.22&0.13&&6.48&3.36&1.72&0.88&0.45&0.24&0.13&\\
    &Error&+0.00&-0.01&+0.00&+0.01&+0.00&-0.01&-0.00&0.01&-0.00&+0.02&-0.03&+0.01&+0.00&-0.01&+0.01&0.01\\
    \hline
    Exponential&Pred.&6.09&3.37&1.21&0.46&0.40&0.40&0.40&&6.35&3.70&1.43&0.52&0.43&0.43&0.43&\\
    &Error&+0.11&-0.31&+0.33&+0.33&+0.01&-0.19&-0.28&0.22&+0.12&-0.32&+0.26&+0.37&+0.03&-0.19&-0.28&0.23\\
    \hline
    Logarithmic&Pred.&4.45&3.55&2.66&1.76&0.87&-0.03&-0.92&&4.74&3.79&2.84&1.90&0.95&0.00&-0.95&\\
    &Error&+1.76&-0.50&-1.12&-0.97&-0.46&+0.24&+1.05&0.87&+1.73&-0.40&-1.15&-1.01&-0.49&+0.23&+1.09&0.87\\
    \hline
    \end{tabular}}
\vspace{1em}
    \centering
    \caption{Comparison of the predicted wall-clock time and its error from the real wall-clock time (VRP)}
    \label{tab:wct_model_comp_vrp}
    \scalebox{0.75}{
    \begin{tabular}{ll||rrrrrrrr!{\bvline{1pt}}rrrrrrrr}
    \hline
    &&\multicolumn{8}{c!{\bvline{1pt}}}{VRP1}&\multicolumn{8}{c}{VRP2}\\
      &&\multicolumn{7}{c}{\# of utilized cores}&&\multicolumn{7}{c}{\# of utilized cores}&\\
        \cline{3-18}
     &&  1&2&4&8&16&32&64&MAE&  1&2&4&8&16&32&64&MAE\\
    \hline
    \rowcolor{gray}Real WCT&&4.04&2.16&1.11&0.60&0.31&0.19&0.15&&10.62&5.59&2.89&1.57&0.82&0.46&0.39&\\
    \hline
    Linear&Pred.&1.86&1.82&1.75&1.60&1.30&0.71&-0.48&&4.86&4.76&4.57&4.18&3.40&1.85&-1.26&\\
    &Error&+2.18&+0.34&-0.64&-1.00&-0.99&-0.52&+0.63&0.90&+5.76&+0.83&-1.67&-2.61&-2.58&-1.38&+1.65&2.35\\
    \hline
    Rational 1&Pred.&4.08&2.09&1.09&0.59&0.34&0.22&0.15&&10.70&5.46&2.84&1.53&0.88&0.55&0.39&\\
    &Error&-0.04&+0.08&+0.02&+0.01&-0.03&-0.03&-0.00&0.03&-0.08&+0.13&+0.05&+0.04&-0.05&-0.09&-0.00&0.06\\
    \hline
    Rational 2&Pred.&4.08&1.18&0.87&0.77&0.73&0.71&0.70&&10.73&3.08&2.27&2.00&1.89&1.84&1.82&\\
    &Error&-0.04&+0.98&+0.23&-0.18&-0.42&-0.52&-0.55&0.42&-0.10&+2.52&+0.63&-0.43&-1.07&-1.38&-1.43&1.08\\
    \hline
    Rational 3&Pred.&4.05&2.15&1.13&0.60&0.33&0.19&0.12&&10.62&5.58&2.92&1.55&0.85&0.50&0.33&\\
    &Error&-0.00&+0.02&-0.02&-0.00&-0.02&-0.01&+0.03&0.01&-0.00&+0.01&-0.02&+0.02&-0.03&-0.04&+0.06&0.03\\
    \hline
    Exponential&Pred.&3.96&2.36&0.95&0.37&0.31&0.31&0.31&&10.41&6.13&2.44&0.96&0.80&0.80&0.80&\\
    &Error&+0.08&-0.20&+0.15&+0.23&+0.01&-0.12&-0.15&0.13&+0.21&-0.54&+0.45&+0.61&+0.02&-0.34&-0.42&0.37\\
    \hline
    Logarithmic&Pred.&2.98&2.40&1.81&1.22&0.64&0.05&-0.54&&7.80&6.27&4.73&3.19&1.66&0.12&-1.42&\\
    &Error&+1.06&-0.23&-0.70&-0.63&-0.32&+0.14&+0.69&0.54&+2.82&-0.67&-1.84&-1.62&-0.83&+0.34&+1.80&1.42\\
    \hline
    \end{tabular}}
\end{table*}
Tables~\ref{tab:wct_model_comp_ppeaks} and \ref{tab:wct_model_comp_vrp} show the comparison of the models for the prediction of wall-clock time. The ``Real WCT'' row shows the median wall-clock time between five different migration gaps in these tables. The rest of the rows indicate the prediction results with the models shown in Appendix~\ref{sec:app_fun}.
The ``Pred.'' row indicates the predicted wall-clock time by each model, while the ``Error'' row indicates the error between the real wall-clock time and the predicted one.
The ``MAE'' column indicates the mean absolute error of each model.

\begin{table}[!tb]
\caption{Rank of the mean absolute error for the prediction of the wall-clock time}
\label{tb:ranking_wct}
\centering
\begin{tabular}{c|c}
\hline
     Rank& Function \\
\hline
     1& Rational 3 \\
     2& Rational 1 \\
     3& Exponential \\
     4& Rational 2 \\
     5& Logarithmic \\
     6& Linear \\
\hline
\end{tabular}
\end{table}
From these results, the Rational 3 model (Eq.~\eqref{eq:rational3_wct}) shows the smallest mean absolute error followed by the Rational 1 model (Eq.~\eqref{eq:rational1_wct}). 
Although the Exponential model also shows a small prediction error, its value is about ten times larger than Rational 3.
All these results show that the mean absolute error is smaller in the order shown in Table~\ref{tb:ranking_wct}.

\begin{figure*}[!tb]
\begin{tabular}{cc}
\begin{minipage}[t]{0.49\linewidth}
    \centering
    \includegraphics[scale=0.6]{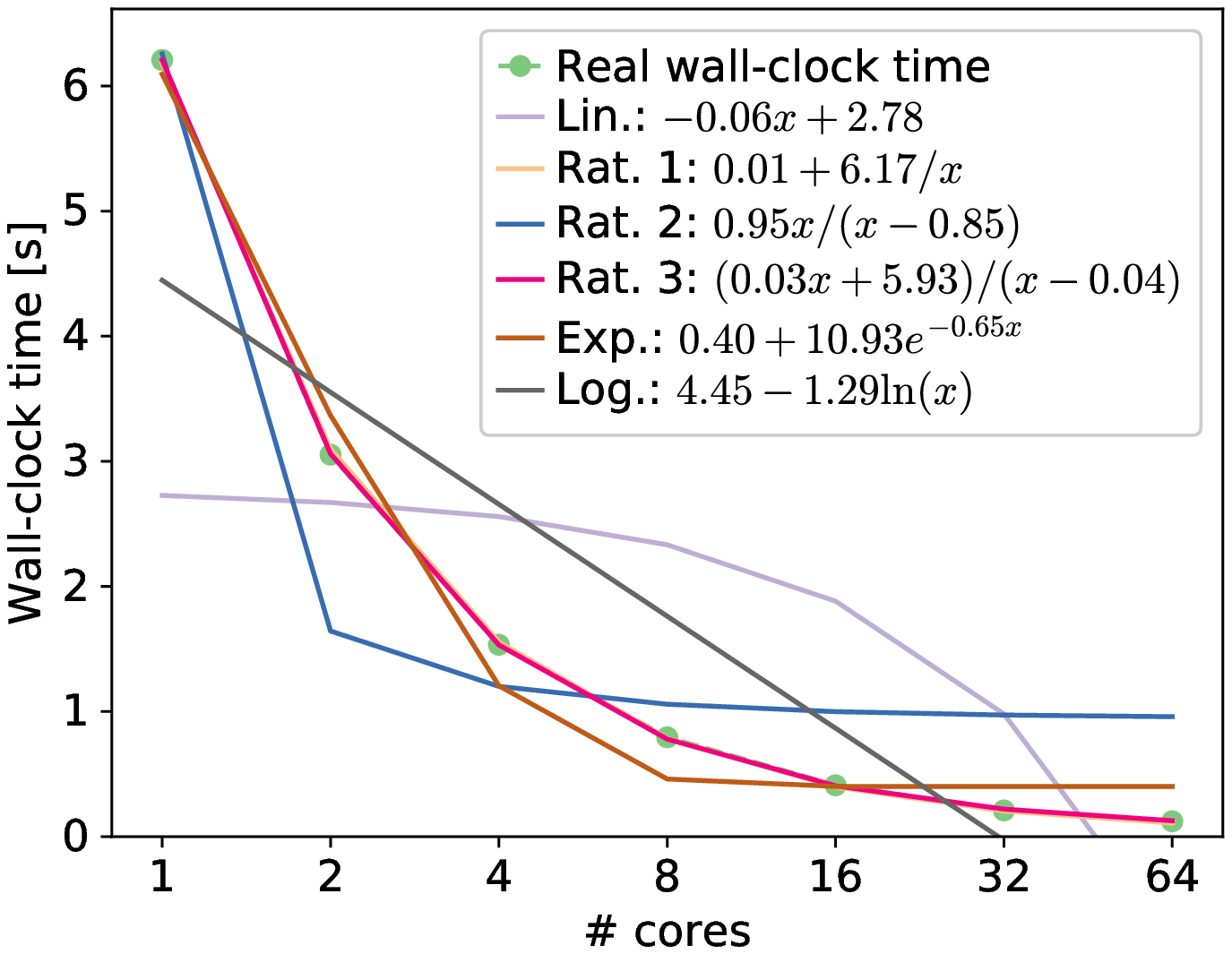}
    \caption{The comparison of mathematical models for predicting the wall-clock time on P-PEAKS 20-100 [s]}
    \label{fig:time_ppeaks_small_comp}
\end{minipage}&
\begin{minipage}[t]{0.49\linewidth}
    \centering
    \includegraphics[scale=0.6]{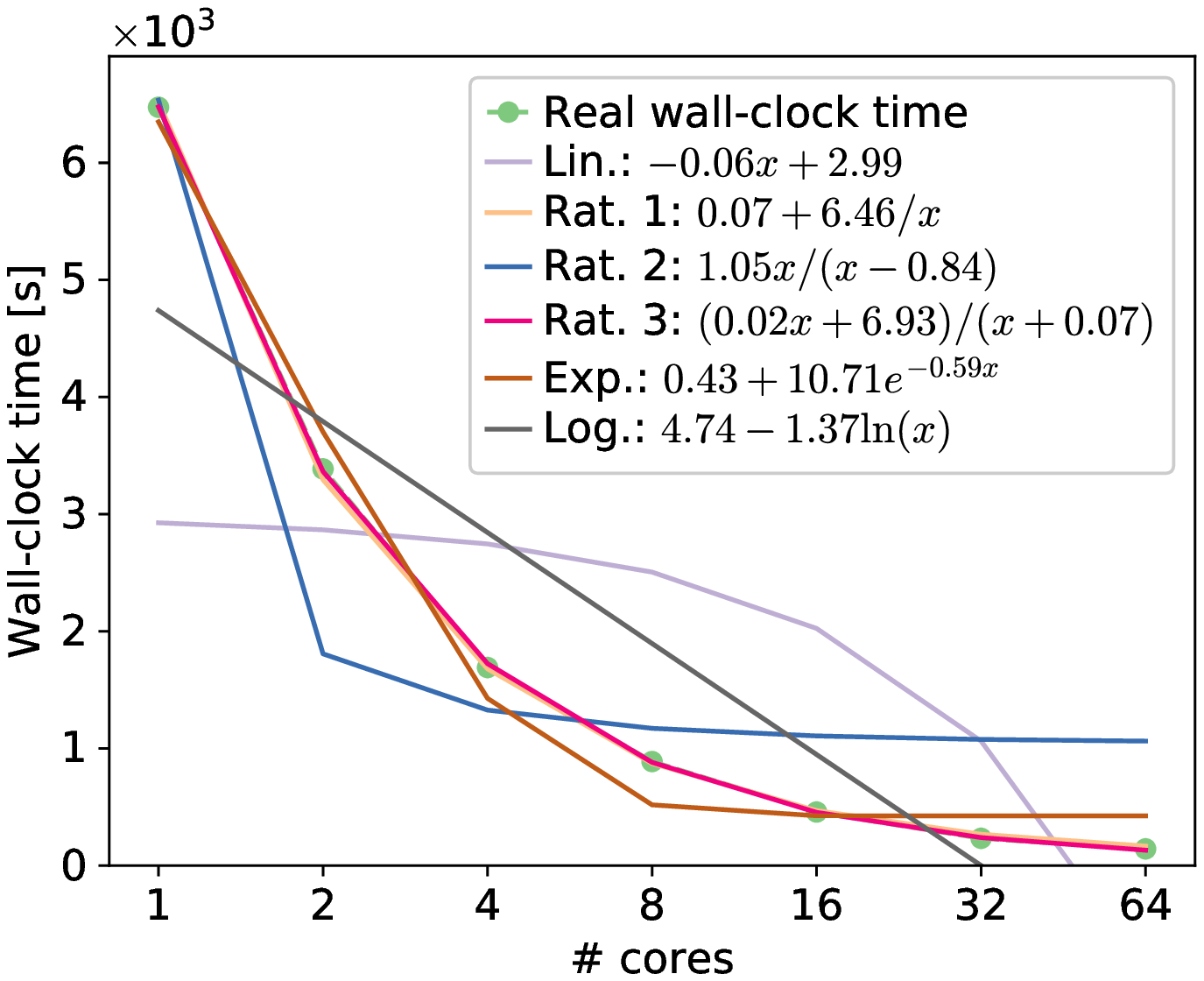}
    \caption{The comparison of mathematical models for predicting the wall-clock time on P-PEAKS 200-1000 [s]}
    \label{fig:time_ppeaks_large_comp}
\end{minipage}\\
\begin{minipage}[t]{0.49\linewidth}
    \centering
    \includegraphics[scale=0.6]{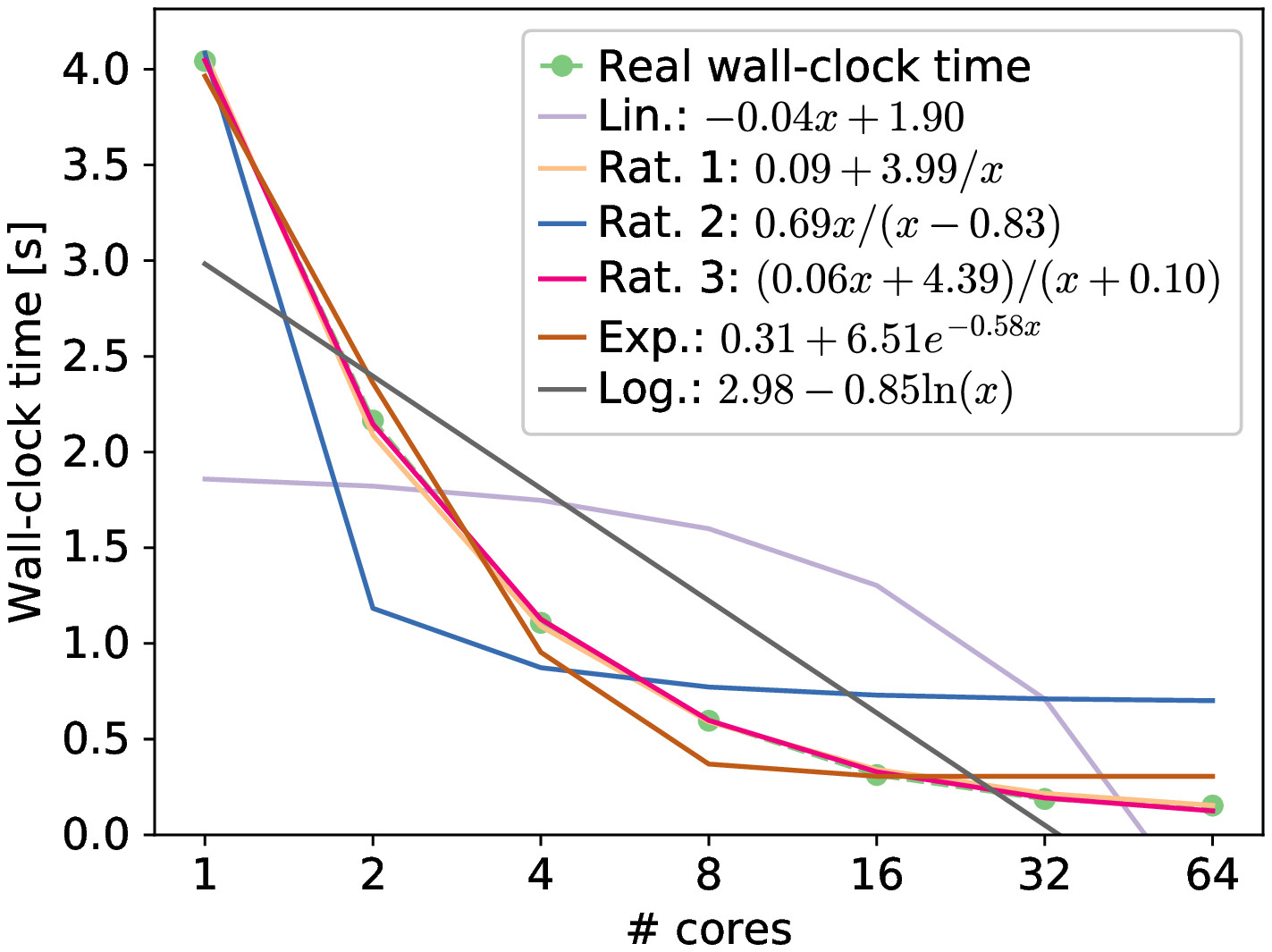}
    \caption{The comparison of mathematical models for predicting the wall-clock time on VRP1 [s]}
    \label{fig:time_vrp1_comp}
\end{minipage}&
\begin{minipage}[t]{0.49\linewidth}
    \centering
    \includegraphics[scale=0.6]{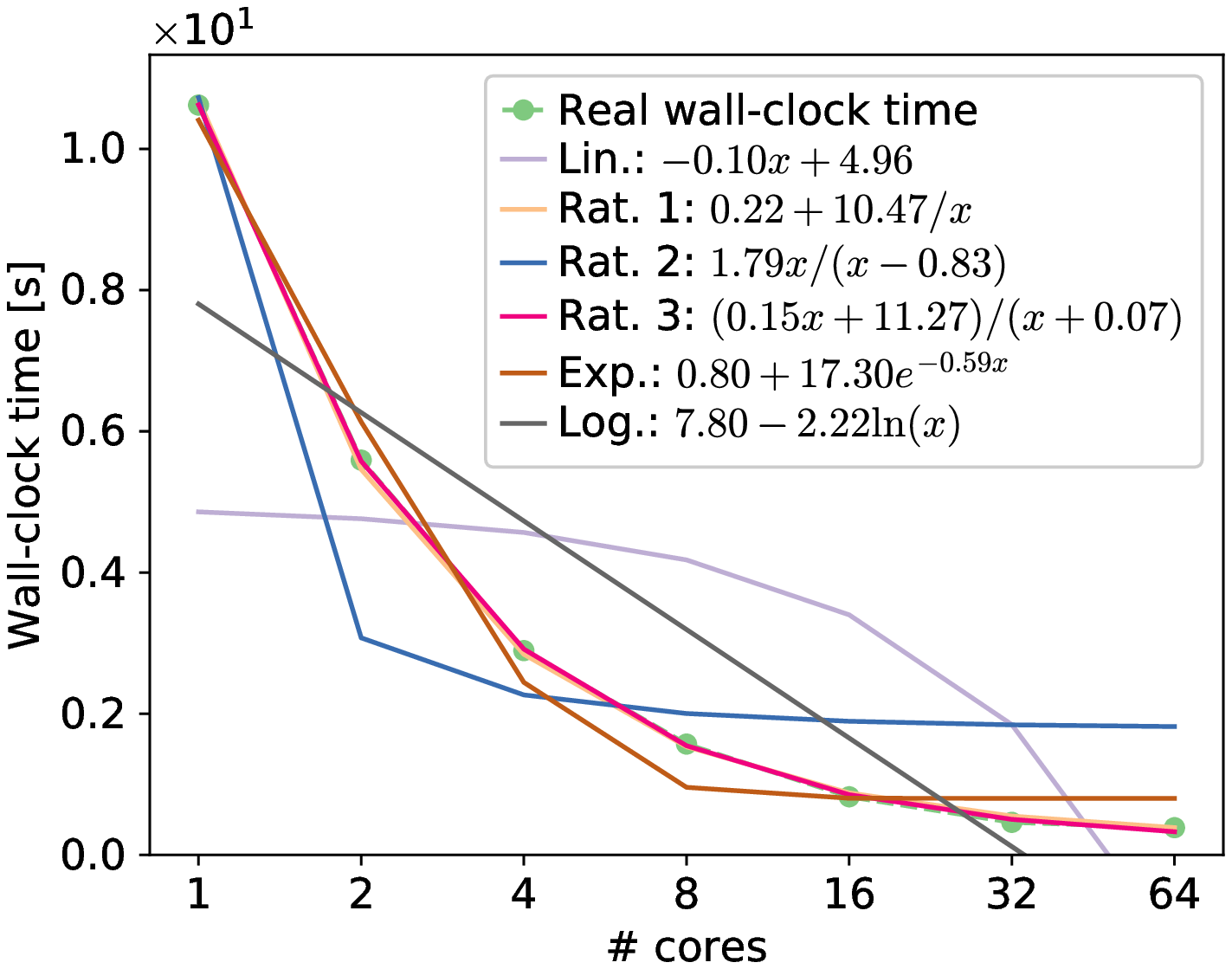}
    \caption{The comparison of mathematical models for predicting the wall-clock time on VRP2 [s]}
    \label{fig:time_vrp2_comp}
\end{minipage}
\end{tabular}
\end{figure*}
Fig.~\ref{fig:time_ppeaks_small_comp}--\ref{fig:time_vrp2_comp} shows the graphical expression of the real wall-clock time and the fitted models.
The horizontal axis shows the number of used cores, while the vertical axis shows the wall-clock time.
The fitted model for each function is denoted in the legend.
Note that the horizontal axis is increasing by a factor of 2.

From these figures, we can see that the Rational 1 and the Rational 3 functions are well fitted to the actual curve of the real wall-clock time. For the other models, the fitted lines are far from the actual line.

From these analyses, it can be said that the Rational 3 function is the most appropriate for expressing the wall-clock time. Therefore, this paper adopted the Rational 3 function for modeling the wall-clock time.

\subsection{Model Selection for the Speed-up}
In this subsection, we show the result of model fitting for the speed-up. As the data for model fitting, we use the speed-up when using the migration gap of 64 shown in Section~\ref{sec:su}.

\begin{table*}[t]
    \centering
    \caption{Comparison of the predicted speed-up and its error from the real speed-up (P-PEAKS)}
    \label{tab:speedup_model_comp_ppeaks}
    \scalebox{0.7}{
    \begin{tabular}{ll||rrrrrrrr!{\bvline{1pt}}rrrrrrrr}
    \hline
    &&\multicolumn{8}{c!{\bvline{1pt}}}{P-PEAKS 20-100}&\multicolumn{8}{c}{P-PEAKS 200-1000}\\
      &&\multicolumn{7}{c}{\# of utilized cores}&&\multicolumn{7}{c}{\# of utilized cores}&\\
        \cline{3-18}
     &&  1&2&4&8&16&32&64&MAE&  1&2&4&8&16&32&64&MAE\\
    \hline
    \rowcolor{gray}Real SU&&1.00&2.06&3.98&7.75&15.18&30.17&52.46&&1.00&1.91&3.86&7.35&14.22&28.05&45.59&\\
    \hline
    Linear&Pred.&1.94&2.76&4.41&7.72&14.32&27.52&53.94&&2.23&2.95&4.39&7.27&13.03&24.54&47.57&\\
    &Error&-0.94&-0.70&-0.43&+0.04&+0.86&+2.65&-1.48&1.01&-1.23&-1.04&-0.54&+0.08&+1.19&+3.51&-1.98&1.37\\
    \hline
    Rational 1&Pred.&-7.36&9.00&17.18&21.27&23.32&24.34&24.85&&-6.44&8.22&15.55&19.21&21.05&21.96&22.42&\\
    &Error&+8.36&-6.94&-13.20&-13.52&-8.14&+5.84&+27.60&11.94&+7.44&-6.31&-11.69&-11.86&-6.83&+6.09&+23.17&10.48\\
    \hline
    Rational 2&Pred.&1.04&2.08&4.12&8.11&15.70&29.50&52.61&&1.03&2.04&4.02&7.84&14.89&27.08&45.84&\\
    &Error&-0.04&-0.02&-0.14&-0.36&-0.52&+0.68&-0.16&0.27&-0.03&-0.13&-0.17&-0.48&-0.67&+0.97&-0.25&0.39\\
    \hline
    Rational 3&Pred.&0.82&1.88&3.97&8.03&15.72&29.61&52.58&&0.71&1.76&3.81&7.73&14.94&27.23&45.79&\\
    &Error&+0.18&+0.18&+0.02&-0.27&-0.54&+0.56&-0.12&0.27&+0.29&+0.15&+0.05&-0.38&-0.72&+0.82&-0.20&0.37\\
    \hline
    Exponential&Pred.&16.09&16.09&16.09&16.09&16.09&16.09&16.09&&14.57&14.57&14.57&14.57&14.57&14.57&14.57&\\
    &Error&-15.09&-14.02&-12.10&-8.33&-0.91&+14.09&+36.37&14.42&-13.57&-12.66&-10.71&-7.21&-0.35&+13.48&+31.02&12.72\\
    \hline
    Logarithmic&Pred.&-7.68&0.25&8.17&16.09&24.01&31.93&39.85&&-6.48&0.54&7.55&14.57&21.58&28.60&35.61&\\
    &Error&+8.68&+1.82&-4.18&-8.33&-8.83&-1.75&+12.60&6.60&+7.48&+1.37&-3.70&-7.21&-7.37&-0.54&+9.98&5.38\\
    \hline
    \end{tabular}}
\vspace{1em}
    \centering
    \caption{Comparison of the predicted speed-up and its error from the real speed-up (VRP)}
    \label{tab:speedup_model_comp_vrp}
    \scalebox{0.7}{
    \begin{tabular}{ll||rrrrrrrr!{\bvline{1pt}}rrrrrrrr}
    \hline
    &&\multicolumn{8}{c!{\bvline{1pt}}}{VRP1}&\multicolumn{8}{c}{VRP2}\\
      &&\multicolumn{7}{c}{\# of utilized cores}&&\multicolumn{7}{c}{\# of utilized cores}&\\
        \cline{3-18}
     &&  1&2&4&8&16&32&64&MAE&  1&2&4&8&16&32&64&MAE\\
    \hline
    \rowcolor{gray}Real SU&&1.00&1.92&3.85&6.89&13.43&21.22&26.10&&1.00&1.91&3.71&6.79&12.92&22.39&27.53&\\
    \hline
    Linear&Pred.&3.63&4.03&4.85&6.49&9.75&16.29&29.36&&3.42&3.86&4.73&6.47&9.96&16.93&30.88&\\
    &Error&-2.63&-2.12&-1.00&+0.40&+3.68&+4.93&-3.27&2.58&-2.42&-1.95&-1.02&+0.32&+2.96&+5.46&-3.35&2.50\\
    \hline
    Rational 1&Pred.&-3.69&6.30&11.30&13.80&15.05&15.67&15.98&&-3.93&6.41&11.59&14.17&15.47&16.11&16.44&\\
    &Error&+4.69&-4.39&-7.45&-6.91&-1.61&+5.55&+10.12&5.82&+4.93&-4.51&-7.88&-7.38&-2.54&+6.28&+11.09&6.37\\
    \hline
    Rational 2&Pred.&1.18&2.30&4.35&7.86&13.19&19.93&26.77&&1.14&2.22&4.24&7.75&13.25&20.54&28.34&\\
    &Error&-0.18&-0.38&-0.50&-0.98&+0.25&+1.30&-0.68&0.61&-0.14&-0.32&-0.53&-0.96&-0.33&+1.85&-0.81&0.70\\
    \hline
    Rational 3&Pred.&0.54&1.79&4.05&7.85&13.40&20.14&26.64&&0.50&1.70&3.91&7.70&13.45&20.77&28.22&\\
    &Error&+0.46&+0.13&-0.20&-0.96&+0.03&+1.09&-0.54&0.49&+0.50&+0.20&-0.20&-0.91&-0.53&+1.63&-0.69&0.67\\
    \hline
    Exponential&Pred.&10.63&10.63&10.63&10.63&10.63&10.63&10.63&&10.89&10.89&10.89&10.89&10.89&10.89&10.89&\\
    &Error&-9.63&-8.71&-6.78&-3.74&+2.80&+10.59&+15.47&8.25&-9.89&-8.99&-7.18&-4.10&+2.03&+11.50&+16.63&8.62\\
    \hline
    Logarithmic&Pred.&-2.60&1.81&6.22&10.63&15.04&19.45&23.86&&-3.01&1.62&6.26&10.89&15.53&20.16&24.80&\\
    &Error&+3.60&+0.11&-2.37&-3.74&-1.61&+1.77&+2.24&2.21&+4.01&+0.28&-2.55&-4.10&-2.60&+2.23&+2.73&2.64\\
    \hline
    \end{tabular}}
\end{table*}
Tables~\ref{tab:speedup_model_comp_ppeaks} and \ref{tab:speedup_model_comp_vrp} show the comparison of models for the prediction of the speed-up. In these tables, the ``Real SU'' row shows the speed-up with the migration gap of 64. The rest of the rows indicate the prediction results with the models shown in Appendix~\ref{sec:app_fun}.
The ``Pred.'' row indicates the predicted speed-up by each model, while the ``Error'' row indicates the error between the real speed-up and the predicted one.
The ``MAE'' column indicates the mean absolute error of each model.

\begin{table}[!tb]
\caption{Rank of the mean absolute error for the prediction of the speed-up}
\label{tb:ranking_speedup}
\centering
\begin{tabular}{c|c}
\hline
     Rank& Function \\
\hline
     1& Rational 3 \\
     2& Rational 2 \\
     3& Linear \\
     4& Logarithmic \\
     5& Rational 1 \\
     6& Exponential \\
\hline
\end{tabular}
\end{table}
From these results, the Rational 3 model (Eq.~\eqref{eq:rational3_wct}) shows the smallest mean absolute error followed by the Rational 2 model (Eq.~\eqref{eq:rational2_wct}). 
The Linear model (Eq.~\eqref{eq:linear}) shows a small prediction error in the P-PEAKS problems while it obtains a larger error in VRPs.
The remaining models produce large prediction errors.
All these results show that the mean absolute error is smaller, in the order shown in Table~\ref{tb:ranking_speedup}.

\begin{figure*}[!tb]
\begin{tabular}{cc}
\begin{minipage}[t]{0.49\linewidth}
    \centering
    \includegraphics[scale=0.6]{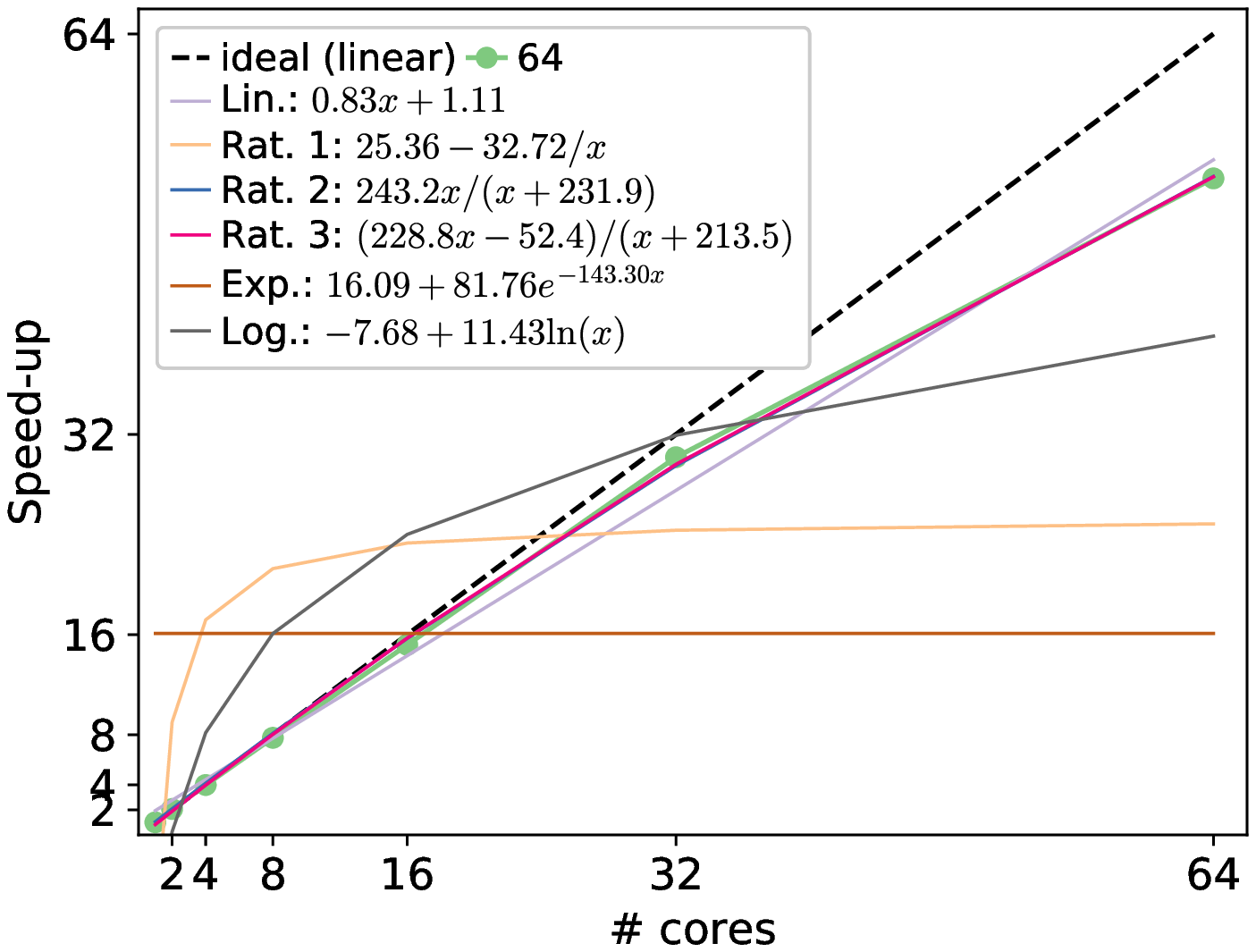}
    \caption{The comparison of mathematical models for predicting the speed-up on P-PEAKS 20-100}
    \label{fig:speedup_ppeaks_small_comp}
\end{minipage}&
\begin{minipage}[t]{0.49\linewidth}
    \centering
    \includegraphics[scale=0.6]{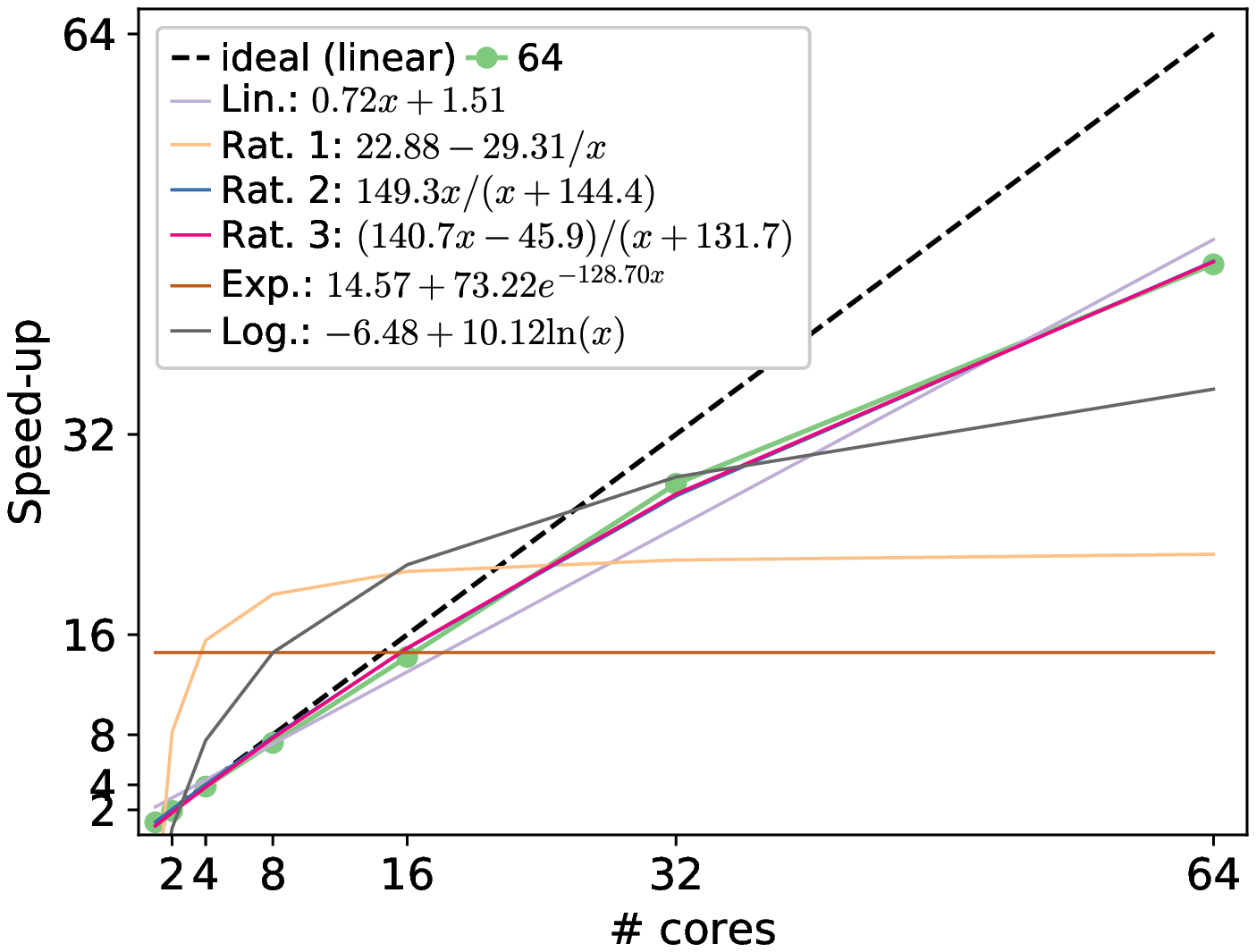}
    \caption{The comparison of mathematical models for predicting the speed-up on P-PEAKS 200-1000}
    \label{fig:speedup_ppeaks_large_comp}
\end{minipage}\\
\begin{minipage}[t]{0.49\linewidth}
    \centering
    \includegraphics[scale=0.6]{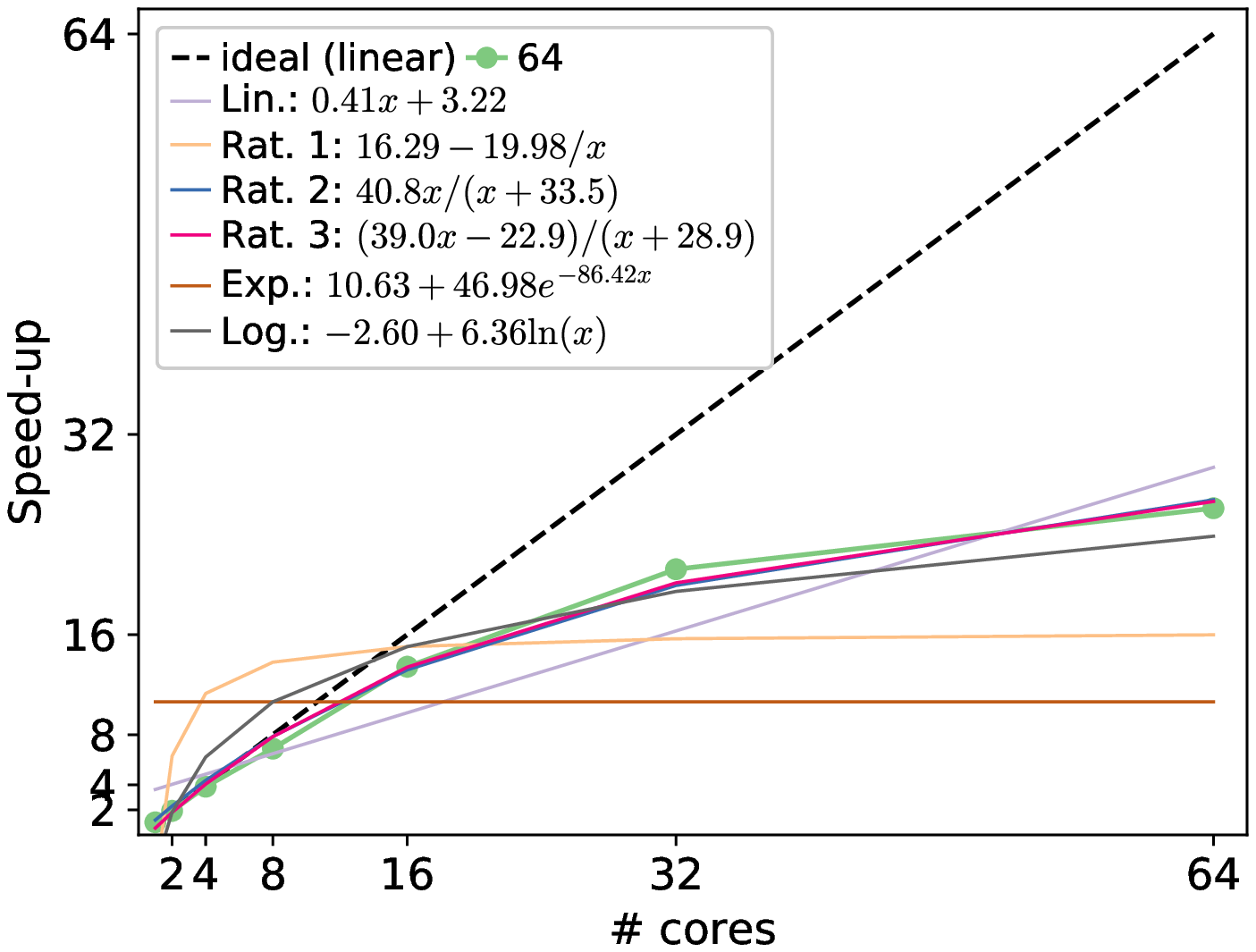}
    \caption{The comparison of mathematical models for predicting the speed-up on VRP1}
    \label{fig:speedup_vrp1_comp}
\end{minipage}&
\begin{minipage}[t]{0.49\linewidth}
    \centering
    \includegraphics[scale=0.6]{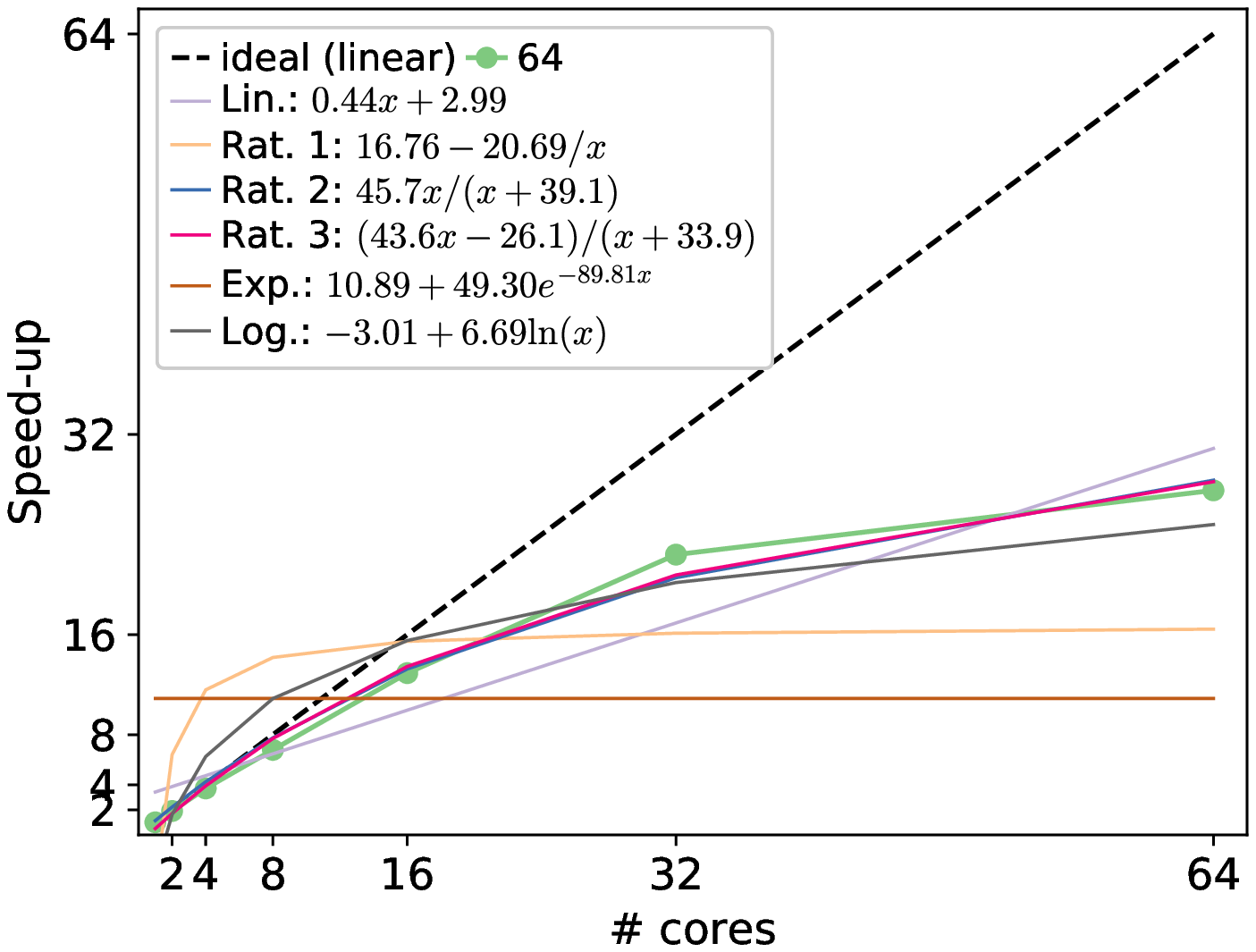}
    \caption{The comparison of mathematical models for predicting the speed-up on VRP2}
    \label{fig:speedup_vrp2_comp}
\end{minipage}
\end{tabular}
\end{figure*}
Fig.~\ref{fig:speedup_ppeaks_small_comp}--\ref{fig:speedup_vrp2_comp} shows the graphical expression of the real speed-up and the fitted models.
The horizontal axis shows the number of used cores, while the vertical axis shows the speed-up.
The fitted model for each function is denoted in the legend.

From these figures, the Rational 2 and the Rational 3 functions both well-fitted to the real speed-up functions for all problems, and they obtain a similar prediction curve.
On the other hand, the Linear function looks suited for the P-PEAKS problems, but it does not reproduce well the curve of VRPs. In contrast, the Logarithmic function seems well-fitting  VRPs, but is not suited for the P-PEAKS problems.
For the remaining functions, the Rational 1 and the Exponential functions, the fitted curves are far from the actual speed-up and are not suitable for approximating the speed-up.

From these analyses, it can be said that the Rational 3 function is the most appropriate for expressing the speed-up. Therefore, this paper adopted the Rational 3 function for modeling the speed-up.

This choice is also consistent with the choice of the Rational 3 function for modeling the wall-clock time.
This is because if the wall-clock time is modeled as $\hat{T}(x)=(ax+b)/(x+c)$, the speed-up is calculated as $\hat{S}(x)=\hat{T}(1)/\hat{T}(x)\equiv (a^\prime x+b^\prime)/(x+c^\prime)$. Therefore, this choice makes sense.

\section*{Acknowledgments}
This research has been supported by Japan Society for the Promotion of Science Grant-in-Aid for Young Scientists Grant Number JP19K20362. Moreover, this research has been partially funded by the European project TAILOR (EC G.A. 952215), and by PRECOG (UMA18-FEDERJA-003), Andaluc\'ia Tech, Universidad de M\'alaga.

\bibliography{references}

\end{document}